\documentclass[sigconf]{acmart}
\AtBeginDocument{%
  \providecommand\BibTeX{{%
    \normalfont B\kern-0.5em{\scshape i\kern-0.25em b}\kern-0.8em\TeX}}}
    
\copyrightyear{2021}
\acmYear{2021}
\setcopyright{rightsretained}
\acmConference[CIKM '21]{Proceedings of the 30th ACM International Conference on Information and Knowledge Management}{November 1--5, 2021}{Virtual Event, QLD, Australia}
\acmBooktitle{Proceedings of the 30th ACM International Conference on Information and Knowledge Management (CIKM '21), November 1--5, 2021, Virtual Event, QLD, Australia}
\acmPrice{15.00}
\acmDOI{10.1145/3459637.3482380}
\acmISBN{978-1-4503-8446-9/21/11}

\usepackage[utf8]{inputenc} % allow utf-8 input
\usepackage[T1]{fontenc}    % use 8-bit T1 fonts
\usepackage{url}            % simple URL typesetting
\usepackage{booktabs}       % professional-quality tables
\usepackage{amsfonts}       % blackboard math symbols
\usepackage{nicefrac}       % compact symbols for 1/2, etc.
\usepackage{microtype}      % microtypography
\usepackage{stmaryrd}
\usepackage{graphicx}
\usepackage{booktabs}       % professional-quality tables

\usepackage{amsfonts,amsmath,amssymb,amsthm}       % blackboard math symbols
\usepackage{nicefrac}       % compact symbols for 1/2, etc.
\usepackage{microtype}      % microtypography
\usepackage{mathtools}
\usepackage{cases}
\usepackage{comment}
\usepackage{paralist}
\usepackage{caption,subcaption}
\usepackage{lipsum}
\usepackage{algpseudocode}
\usepackage{breqn}
\usepackage{algorithm}
\usepackage{color}
\usepackage{appendix}
\usepackage{xspace}
\usepackage{enumitem}
\usepackage{gensymb}
\usepackage[english]{babel}
\usepackage{xcolor}
\usepackage{wrapfig}
 \usepackage{multirow}
 \usepackage{graphicx}
 \usepackage{diagbox}
 \usepackage{bm}

\mathchardef\mhyphen="2D

\newcommand{\vertiii}[1]{{\left\vert\kern-0.25ex\left\vert\kern-0.25ex\left\vert #1
    \right\vert\kern-0.25ex\right\vert\kern-0.25ex\right\vert}}

\def\bd{{\mathbf{d}}}

\def\bu{{\mathbf{u}}}
\def\bv{{\mathbf{v}}}

\def\bx{{\mathbf{x}}}

\def\bz{{\mathbf{z}}}
\def\bD{{\mathbf{D}}}

\def\bO{{\mathbf{O}}}

\def\bU{{\mathbf{U}}}
\def\bV{{\mathbf{V}}}
\def\bW{{\mathbf{W}}}
\def\bX{{\mathbf{X}}}
\def\bY{{\mathbf{Y}}}
\def\bZ{{\mathbf{Z}}}

\def\cG{\mathcal{G}}

\newtheorem{theorem}{Theorem}

\DeclarePairedDelimiterX{\infdivx}[2]{(}{)}{%
  #1\;\delimsize\|\;#2%
}

\newcommand\blfootnote[1]{%
  \begingroup
  \renewcommand\thefootnote{}\footnote{#1}%
  \addtocounter{footnote}{-1}%
  \endgroup
}

\definecolor{alggreen}{RGB}{36,135,74}
\definecolor{algblue}{RGB}{29,138,168}
\definecolor{algorange}{RGB}{125,75,0}

\begin{document}
\fancyhead{}
\title{Pulling Up by the Causal Bootstraps: Causal Data Augmentation for Pre-training Debiasing}

\author{Sindhu C. M. Gowda}
\email{sindhu.gowda@mail.utoronto.ca}
\affiliation{%
  \institution{University of Toronto}
  \institution{Vector Institute}
  \city{Toronto}
  \state{Ontario}
  \country{Canada}
}

\author{Shalmali Joshi}
\email{shalmali@seas.harvard.edu}
\affiliation{%
  \institution{Harvard University}
  \city{Cambridge}
  \state{Massachusetts}
  \country{USA}
  }

\author{Haoran Zhang}
\email{haoran@cs.toronto.edu}
\affiliation{%
  \institution{University of Toronto}
  \institution{Vector Institute}
  \city{Toronto}
  \state{Ontario}
  \country{Canada}
}

\author{Marzyeh Ghassemi}
\email{mghassem@mit.edu}
\affiliation{%
  \institution{MIT}
  \city{Cambridge}
  \state{Massachusetts}
  \country{USA}
}

\renewcommand{\shortauthors}{Gowda et al.}

\begin{abstract}
Machine learning models achieve state-of-the-art performance on many supervised learning tasks. However, prior evidence suggests that these models may learn to rely on “shortcut” biases or spurious correlations (intuitively, correlations that do not hold in the test as they hold in train) for good predictive performance. Such models cannot be trusted in deployment environments to provide accurate predictions. While viewing the problem from a causal lens is known to be useful, the seamless integration of causation techniques into machine learning pipelines remains cumbersome and expensive. In this work, we study and extend a causal pre-training debiasing technique called causal bootstrapping (CB) under five practical confounded-data generation-acquisition scenarios (with known and unknown confounding). Under these settings, we systematically investigate the effect of confounding bias on deep learning model performance, demonstrating their propensity to rely on shortcut biases when these biases are not properly accounted for. We demonstrate that such a causal pre-training technique can significantly outperform existing base practices to mitigate confounding bias on real-world domain generalization benchmarking tasks. This systematic investigation underlines the importance of accounting for the underlying data-generating mechanisms and fortifying data-preprocessing pipelines with a causal framework to develop methods robust to confounding biases.

% Machine learning models achieve state of the art performance on many supervised learning tasks. However, prior evidence suggests these models may learn to rely on ``shortcut'' biases or spurious and undesirable correlations for good predictive performance. Such models cannot be trusted in deployment settings to provide accurate predictions. There is a signficant gap in knowledge of whether pre-training debiasing methods can address this challenge. In this work, we study and extend pre-training debiasing procedure using five practical confounded-data generation-acquisition scenarios (with known and unknown confounding).  
% Under these settings, we systematically investigate the effect of confounding bias on deep learning model performance, demonstrating their propensity to rely on shortcut biases when these biases are not properly accounted for. We extend a causal sampling technique, called causal bootstrapping, to demonstrate that such a pre-training technique can significantly outperform existing methods to mitigate confounding bias on real-world domain generalization benchmarking tasks.
% This systematic investigation underlines the importance of accounting for the underlying data-generating mechanisms and fortifying data-preprocessing pipelines with a causal framework to develop methods robust to confounding biases.
\end{abstract}

\begin{CCSXML}
<ccs2012>
<concept>
<concept_id>10010147.10010257</concept_id>
<concept_desc>Computing methodologies~Machine learning</concept_desc>
<concept_significance>500</concept_significance>
</concept>
</ccs2012>
\end{CCSXML}

\ccsdesc[500]{Computing methodologies~Machine learning}

\keywords{confounding bias, pre-training, debiasing, causal graphs, re-sampling}

\maketitle

\section{Introduction}
\label{sec:intro}
Machine learning (ML) models have achieved state-of-the-art performance on safety-critical tasks ranging from self-driving cars \cite{bojarski2016end} to disease prediction \cite{gulshan2016development,esteva2017dermatologist}. In many real settings, models are found to rely on specific biases present in their training environment as ``shortcuts'' for successful prediction ~\cite{jo2017measuring,beery2018recognition, geirhos2018imagenet}. 
For instance,~\citet{zech2018variable} found that deep learning models exploited chest X-ray data's hospital of origin, rather than disease-specific features, to detect pneumonia.~\citet{zhang2020hurtful} showed that language models capture relationships between gender and medical conditions which exceed biological associations. Other studies have uncovered a worrisome reliance on gender in models trained for
recommending jobs \cite{dastin2018amazon}, and race in prioritizing patients for medical care in spite of comparable risk~\cite{obermeyer2019dissecting}. 
In these cases, the hospital of origin or person's gender and race act as the source of ``shortcut'' bias respectively. This is known as \emph{confounding bias} and occurs when some attributes are systematically correlated to the prediction label and data features. A naively trained model learns to rely on the \emph{confounding} biases rather than the true association between data features and outcomes. This effectively obscures a model's ability to learn the true relationship between data and outcome~\cite{pearl2009causality,hernanchapman}. When deployed in an environment where these spurious associations change or no longer exist, the model performs poorly. \blfootnote{Code: \url{https://github.com/MLforHealth/CausalDA}}
While understanding the problem from the lens of causation can be powerful, most causal techniques (e.g. \citet{peters2016causal, subbaswamy2019preventing}) scale poorly with respect to the number
of variables in the learning problem in terms of computational complexity. As such, the integration of causation tools into the machine learning pipeline remains cumbersome and expensive. Further, to reliably debias models, a careful augmentation to the training pipeline is required. 
% , either in the form of assumptions on the underlying generative process or assumptions about deployment environments. 
Existing work has primarily sought to address this challenge using expensive model specific \emph{training} procedures \cite{kim2019learning,wang2019balanced,zemel2013learning,quadrianto2019discovering,bahng2020learning}. On the other hand,
pre-training methods for debiasing have received relatively little attention in the literature~\cite{chyzhyk2018controlling,landeiro2017controlling}. 
One baseline pre-training method is to use domain knowledge to select features we think are predictive of the label and want the model to learn upon. However, this does not explicitly account for possible biases, e.g., chest x-ray data to predict disease labels without accounting for possible hospital metadata biases \cite{zech2018variable}. Others resort to utilizing as many features as possible to rely on models themselves to filter out spurious correlations~\cite{suresh2017clinical,ghassemi2014unfolding}. In some cases, we explicitly account for confounding, by upsampling data w.r.t the confounding to ensure invariances to confounding bias (simple data augmentation) ~\cite{sharma2020data,panda2018contemplating,geirhos2018imagenet,zhang2020towards,goel2020model}. Such information is not always explicit, resulting in bias mis-specification~\cite{geirhos2018imagenet,wang2019balanced}.
To address these problems, we study causal bootstrapping (CB) -- data re-sampling based on causal information -- as proposed by \citet{little2019causal} which uses prior knowledge of the data generating process to debias models. Their utility however has only been tested for simple data generation scenarios and evaluated on simple model classes, such as random forest and logistic regression. To the best of our knowledge, there has been no study analyzing causal pre-training debiasing techniques like CB for learning unbiased deep models.
% in detail.
%\textcolor{red}{Some methods (Simple) train directly on confounded data only using features that they want the model to learn their label based on. 
%}
%Other options use proactive approaches which anticipate and address these biases in training environments through pre-training techniques.    
% Another option is to train on data that is first \emph{de-confounded} through pre-processing. 
%These approaches are commonly known as data augmentation (DA) and used to de-bias training data . 
%However to completely deconfound the data, this procedure involves . 

% Other methods explore training techniques that constrain models to learn without biases by training them to be independent of particular confounding variables \citep{janizek2020adversarial,wang2019balanced,zemel2013learning} or using additional helper models~\citep{bahng2019learning,sharma2020data,peyre2017weakly}. However, these methods either demand knowledge of confounding and/or specific architectures to perform de-biasing. 

In this work, we first perform a systematic analysis to provide evidence that complex deep neural network models are prone to \emph{confounding} biases. 
%and characterize the extent to which this prevents generalization.
% We then explore the viability of specifying the underlying causal generating mechanism and leveraging Causal Bootstrapping as a pre-training debiasing method.
%We focus on CB, because it directly uses the causal generative process. 
We extend CB, by deriving CB weights under more complex and realistic data acquisition scenarios and analyse their debiasing performance. 
% We contrast our extension of CB with existing methods for training confounded models (simple and IF) and de-confounded models (DA). 
We contrast CB with existing popular pre-training methods, including data-augmentation, to demonstrate benefits of a causal perspective in pre-training debiasing without additional training costs.  
%for training models (simple, IF and DA). 
We study these methods under five \emph{acquisition scenarios} motivated by realistic settings: a) observed confounding, b) observed confounding with mediator (a variable that directly influences the input based on the label, and is affected by confounding only through label), c) partially observed confounding with mediator, d) unmeasured confounding with a mediator, e) observed confounding with biased level of care.

We demonstrate our results on synthetic and semi-synthetic data with synthesized and real-world shifts using six datasets: CelebA~\cite{liu2015faceattributes}, ChestXray18~\cite{wang2017chestx}, CheXpert \cite{irvin2019chexpert}, MIMIC-CXR~\cite{johnson2019mimic}, Camelyon17 \cite{bandi2018detection} and PovertyMap~\cite{yeh2020using} from the recent WILDS benchmark~\cite{koh2020wilds}. We demonstrate benefits of proposed pre-training method to train models that generalize to unseen test environments, providing evidence that such methods help deep networks rely on generalizable associations in the data as opposed to spurious ones. 

%We train deep neural models having access to \emph{only} confounded data under each setting using all four methods (Simple, IF, DA, CB). 
% We evaluate them on 3 specifically designed test environments to understand the extent to which each model relies on confounding. 
%\textcolor{red}{To measure model reliance, we generate semi-synthetic confounded benchmark datasets of varying task difficulty, spanning across different image classification tasks to simulate our five confounding scenarios: digit~\cite{lecun1998gradient}, gender ~\cite{liu2015faceattributes}, chest X-ray~\cite{wang2017chestx} and tumor tissue prediction \cite{koh2020wilds}. We further show our results on dataset with both synthetic and real-world data shifts acting as sources of confounding.
%We evaluate the efficacy of models trained with the four methods for each of the five confounding data scenarios on three specifically designed test environments --- confounded test, unconfounded test, and reverse-confounded test --- to check model robustness to confounding. 
Our observations are as follows:
\begin{asparaenum}
% \item[1)] \textcolor{red}{Deep neural networks rely on the confounding bias signals for prediction. While these models can overcome confounding bias when spurious correlations are small, they increasingly rely on spurious confounding as these correlations increase and especially when data-generating mechanisms are also complex.} 
\item[1)] Deep networks trained on selected ``predictive'' features without accounting for biases tend to rely on bias signals for prediction. While this effect is small when correlations are low, the reliance on confounding increases as these correlations increase. This is shown as a drop in performances of up to $40\%$ on test sets with different bias-label correlations for simulated multivariate Gaussian data and up to $50\%$ for real-world medical data.   
% , especially with complex data-generation-acquisition mechanisms. 
% \item[2)] Models trained either ignoring confounding features or naively including without adjusting for them properly perform poorly across all scenarios specifically at high correlations. Naively using all features but fail significantly for high correlations particularly when such features do not include actual confounding features but could contain a mediator variable (defined in Section~\ref{sec:experiments}).https://www.overleaf.com/project/60ad44361944029b2d77ecf3
\item[2)] Models trained with all available features without adjusting for the confounding, perform poorly when spurious correlations are high. We observe this across all five evaluation scenarios in all six datasets. 
% Naively using all features but fail significantly for high correlations particularly when such features do not include actual confounding features but could contain a mediator variable (defined in Section~\ref{sec:experiments}).

\item[3)] While simple data-augmentation performs well when the source of confounding is known, CB is the only method able to leverage causal information to perform well when the confounding is unknown. CB only needs the appropriate causal quantity to be identified.  

\item[4)] Our real-world analyses provides strong evidence that models debiased using CB do not learn on confounding bias,
and hence show similar performance on test data with unknown changes in confounding bias-label correlations.
\end{asparaenum}
% \fixme{needs re-writing}

\section{Related work}
\label{sec:rel_work}
Supervised ML models have been shown to learn on spurious confounding~\cite{buolamwini2018gender,suresh2019framework,zhao2019gender}, with many rigorous solutions recently proposed to tackle this problem~\cite{arjovsky2019invariant,peters2015causal,heinze2018invariant,little2019causal}. We summarize methods used during either model pre-training and in-training.

% \vspace{-2mm}
\subsection{Pre-training Debiasing of Data}
Several methods have been proposed to address biased data prior to model training.~\citet{chyzhyk2018controlling} propose ``anti'' mutual-information sub-sampling to create de-biased samples. However it does not guarantee marginal distributions are retained.
% On the other hand, we focus even on causal methods to remove the dependence between the model predictions and the bias.
~\citet{landeiro2017controlling} use predictive models to learn unobserved confoundings from a few samples with measured confounding. These methods are restricted to either a single measured confounder, or a pre-specified bias for many inputs.  Another approach to remove confounder influence is matching samples to improve data balance. Given the kind of bias, when additional examples can be collected like in~\citet{panda2018contemplating} or supplied through generating functions for images~\cite{geirhos2018imagenet} or text~\cite{shah2019cycle}, the training data itself can be de-confounded. However, building a generative model with pre-defined bias types can easily suffer from bias mis-specification and lacks practicality.
% \textcolor{red}{\citet{teshima2021incorporating} exploit conditional independence assumptions as seen in causal graphs to design augmentation methods to improve performance in small data regime.} 
If specific information about which parts of the causal graph to be invariant to is known,~\citet{subbaswamy2019preventing} and \citet{subbaswamy2020spec} present a causal algorithm for identifying the corresponding stable distribution independent of these biases. Usually specifying this information requires domain expertise. Instead, we focus on removing all sources of confounding. Also, unlike the work on causal transportabily in \citet{pearl2011transportability}, we do not assume access to multiple heterogeneous sources with distinct experiments. We note that \textit{"A Causal bootstrap"} by \citet{imbens2018causal} has a similar name, but proposes to resolve statistical issues in average causal effect estimates, as opposed to our work in confounding bias. 

% that the models can learn based on. 
% A broader approach invokes causal stability, constructing predictors which only exploit specific information that we want the model to learn based on. For example, invariant causal prediction methods combine data from multiple debiased
% samples; using this they select or find a representation of  variables that directly effect the prediction target \citep{peters2015causal,arjovsky2019invariant}. They use this information to construct a predictor which is invariant across differing target settings. 

\subsection{In-training Debiasing of Models}
Several methods improve training procedures to learn models invariant to confounding bias.~\citet{peters2015causal,arjovsky2019invariant,heinze2018invariant} use multiple datasets from various sources to construct predictors that exploit invariant information within them during training. 
% (e.g. chest x-rays from multiple hospital sites to train diagnosis models). 
Other approaches have proposed removing predictability of bias based on input/outcome through domain adversarial losses \citep{wang2019balanced} or mutual information minimization \citep{kim2019learning}. These methods depend on the knowledge of bias for every sample, which is often difficult to enumerate. A few in-training approaches explored in the context of algorithmic fairness, encourage independence to the sensitive attribute ~\citep{zemel2013learning,quadrianto2019discovering}.
%Another set of approaches deal with in-processing to prevent models from learning on confounding bias, where new loss functions are explored to implicitly enforce fairness -- proposing independence-based regularisers to encourage independence between the bias and the prediction \citep{zemel2013learning} or the conditional independence given the outcome \citep{quadrianto2019discovering}. 
While, \citet{bahng2020learning,zhang2020causal} propose methods where explicit knowledge of confounding is not essential, they still require a bias-characterising model. 
% to generate debiased predictions. 
% While these models hugely benefit from data augmentation, they have only been tried on smaller architectures. 
When explicit knowledge of bias is unavailable, but data from a specific target environment is, domain adaptation has been used to re-weight samples from the source for their likelihood on the target~\citep{rao2017predictive, ben2007analysis, cortes2014domain,byrd2019effect,cortes2010learning}. 
In contrast, we address known and unknown confounding biases without access to target data. %, and contextualize the effectiveness of various pre-processing methods for training models robust to confounding bias.

%~\citet{rao2017predictive} discuss instance reweighting, where samples are re-weighted in the training set to minimize the empirical risk in the target population instead of the source population. This method assumes access to a target test set, which we do not assume. 

\label{sec:methods}
\section{Methods}\label{sec:methods_all}
To explore the effect of pre-training debiasing on deep networks, we use causal machinery to design our framework. We then demonstrate how pre-training methods rely on spurious correlations. Next, we extend causal bootstrapping to general data-acquisition mechanisms (under certain conditions) and show its debiasing capabilities. 
We first introduce notations and preliminaries.

\begin{figure*}[t!]
    \centering
    \begin{subfigure}[t]{.19\textwidth}
     % \centering
      \includegraphics[width=1.5\textwidth, trim=250 280 100 220, clip]{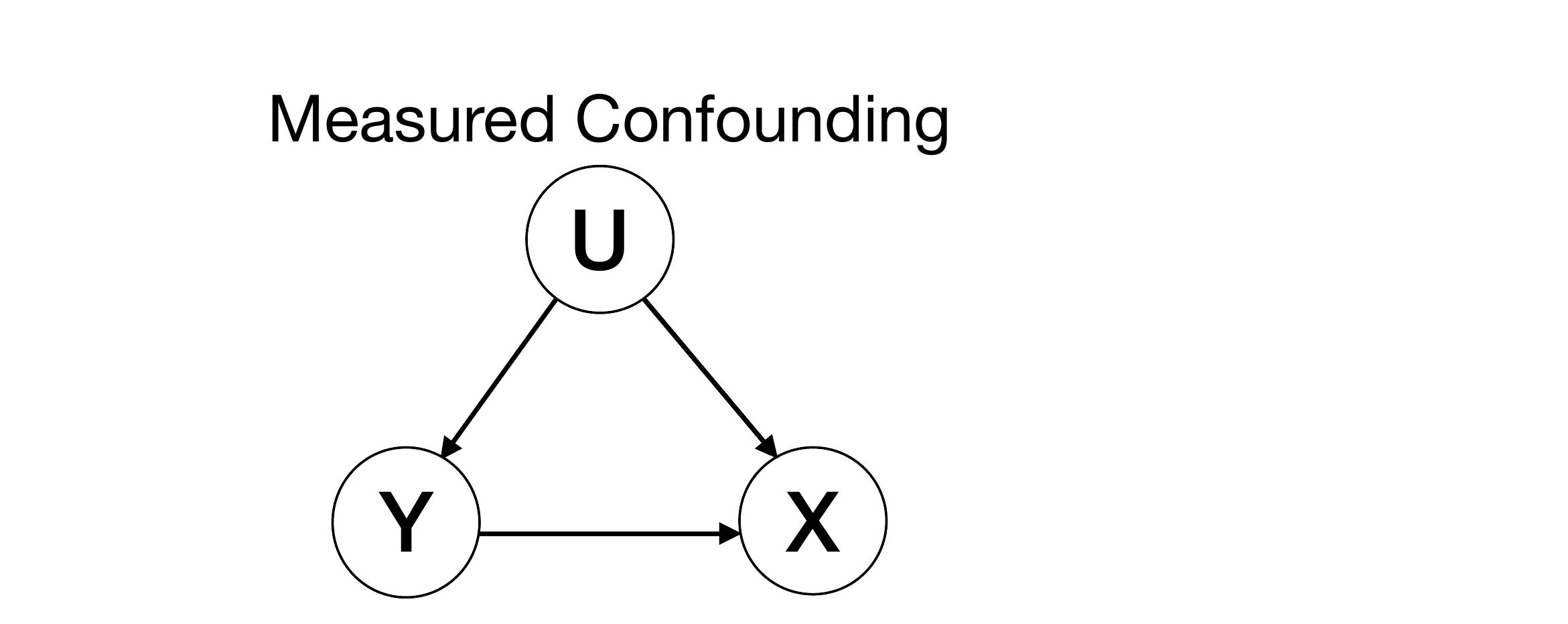}
      \caption {Observed confounding}
      \label{fig:mes_con}
    \end{subfigure}
    \begin{subfigure}[t]{.19\textwidth}
      \centering
      \includegraphics[width=1.5\textwidth, trim=250 280 100 220, clip]{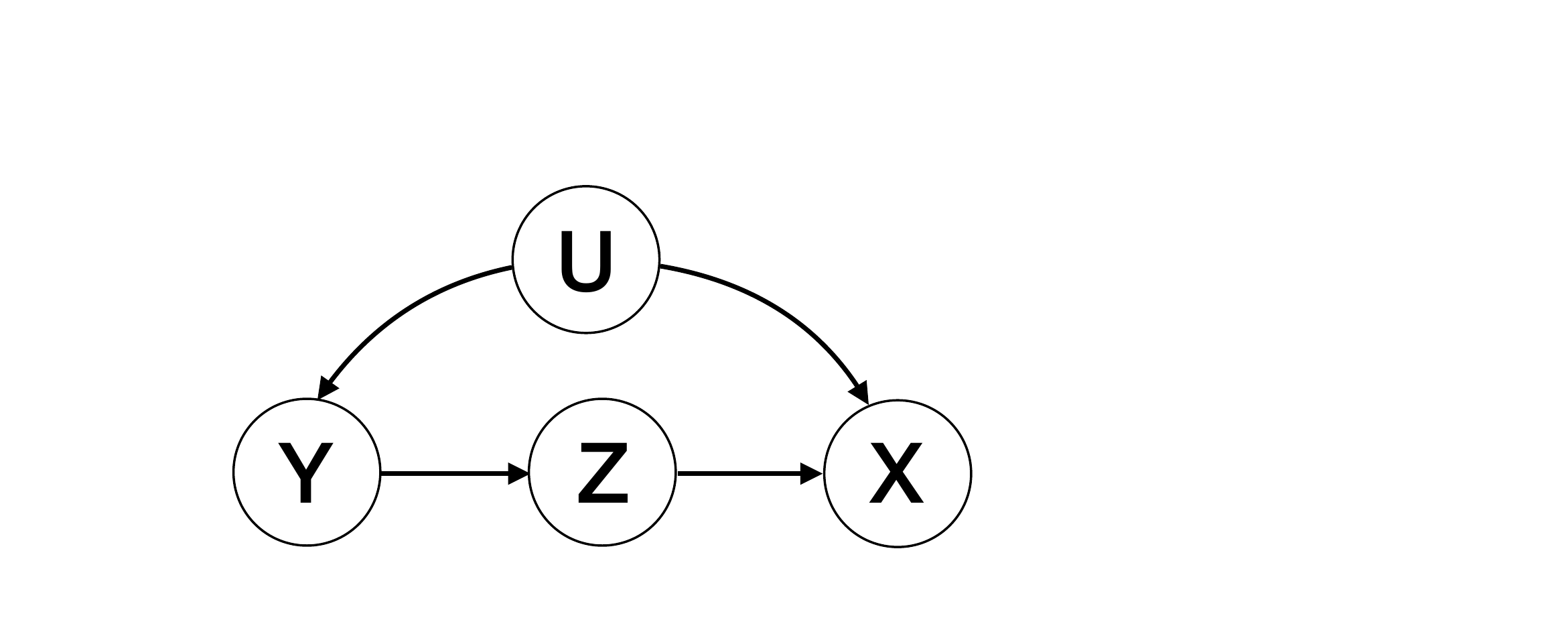}
      \caption {Observed confounding with mediator}
      \label{fig:mes_wm}
    \end{subfigure}
    \begin{subfigure}[t]{.19\textwidth}
      \centering          
      \includegraphics[width=1.5\textwidth, trim=250 280 100 220, clip]{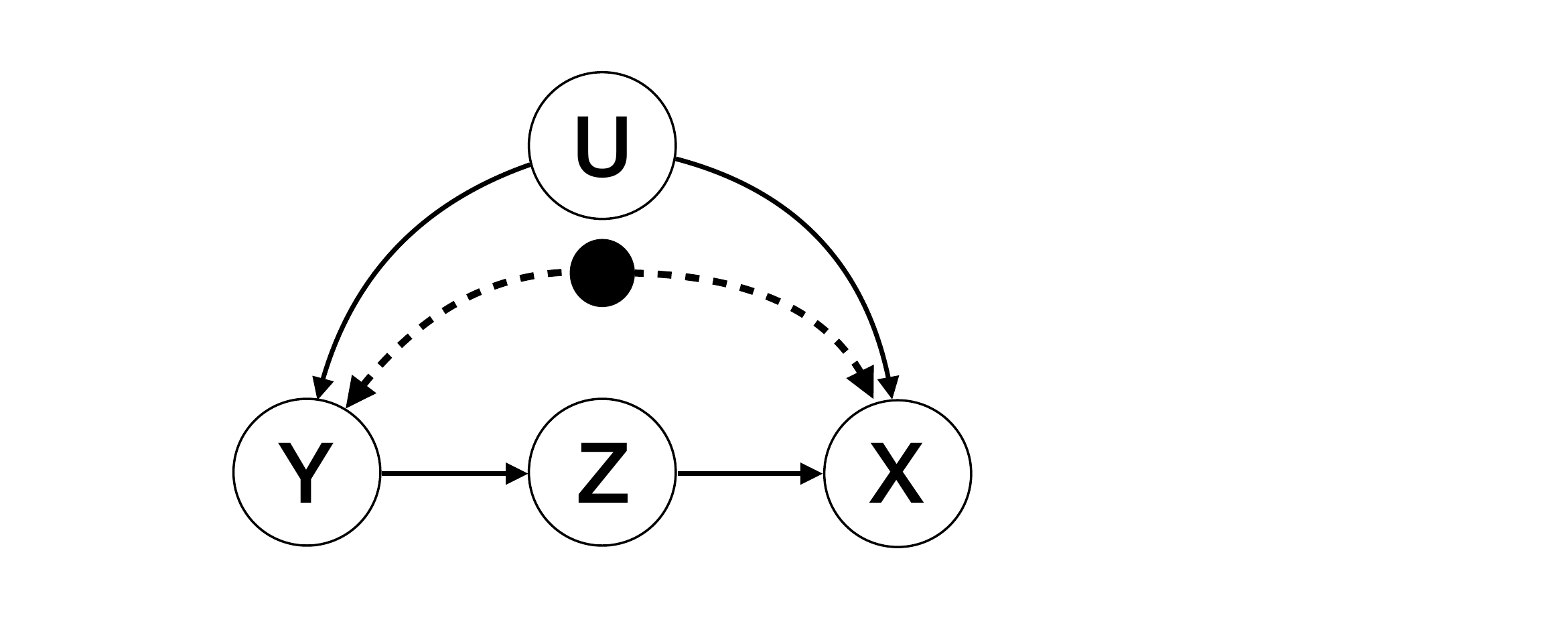}
      \caption {Partially observed confounding with mediator}
      \label{fig:pmes_wm}
    \end{subfigure}
    \begin{subfigure}[t]{.19\textwidth}
      \centering          
      \includegraphics[width=1.5\textwidth, trim=250 280 100 220, clip]{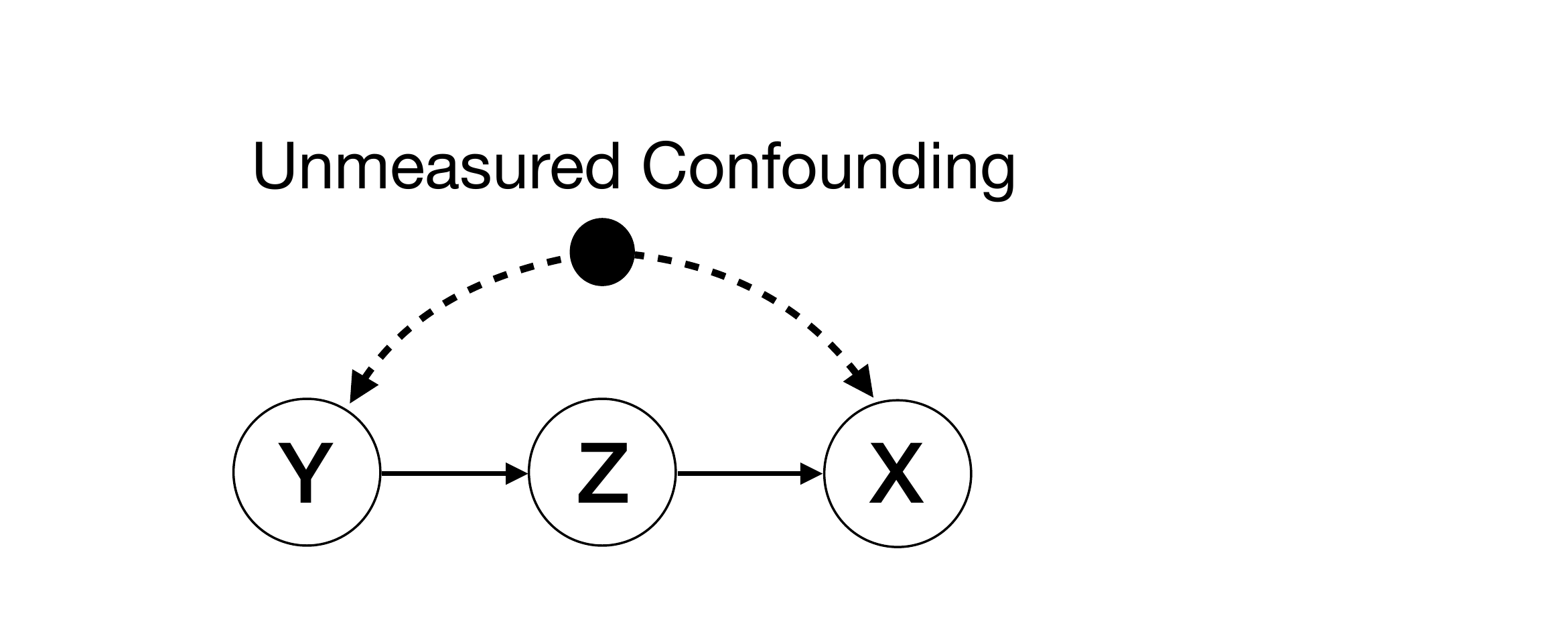}
    %   {img/obs_mediator.pdf}
      \caption {Unobserved confounding with mediator}
      \label{fig:unmes_wm}
    \end{subfigure}
     \begin{subfigure}[t]{.19\textwidth}
     % \centering
      \includegraphics[width=1.5\textwidth, trim=250 280 100 220, clip]{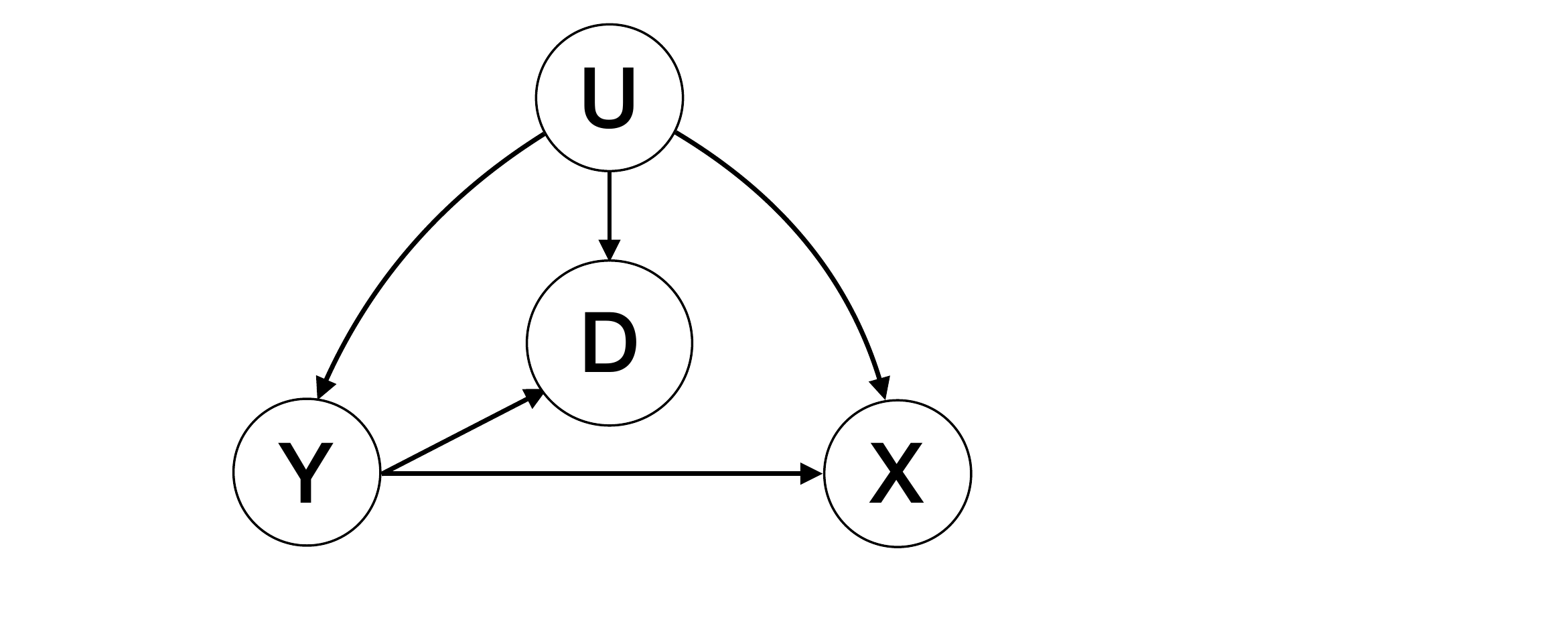}
      \caption {Biased care with Observed confounding}
      \label{fig:mes_wom}
    \end{subfigure}
    \caption{Data acquisition settings. Example: $Y$: disease, $\bX$: symptoms, $\bU$: hospital of acquisition or race (observed confounding), $\bZ$: disease bio-marker information (mediator), $D$: level of care.}
    \label{fig:DAG}
\end{figure*}

\paragraph{\textbf{Notations and preliminaries.}} 
Capital letters $X$, $Y$, $Z$, $D$ and $U$ represent random variables while $x$, $y$, $z$, $d$ and $u$ are their realizations. Multi-dimensional random variables and realizations are in bold i.e. $\bX$ and $\bx$. We represent data acquisition scenarios using causal graphs (Fig. \ref{fig:DAG}). In a causal graph, nodes represent (observed or unobserved) variables of interest and directed edges represent their causal dependence~\citep{pearl2009causality}. Variables with a bidirectional (dotted) edge have a latent (unknown) common cause which acts as confounding, shown as a black node. Usually, $\mathbf{X}, \bZ, \bD, \bU$ represent covariates. Note that they are not always collected i.e. we only know of their presence. $Y$ denotes the target label to be learned.
{\small{
\begin{table}[thbp!]
    \begin{tabular}{p{0.16\textwidth}|p{.28\textwidth}}
    \hline
    \bf{Term} & \bf{Definition}\\ 
    \hline
    Confounded data & Data samples from a process that has confounding bias. (e.g. processes in Figure~\ref{fig:DAG} are all confounded)\\ \hline
    Unconfounded data & Data samples from a process that does not have confounding bias.\\ 
    \hline
    Deconfounded data & Data samples from a process that has confounding bias, but is debiased by a\\ 
                      &  pre-training method.\\ 
    \hline
    \end{tabular}
\caption{Common terms used in the paper}
\label{tab:terms}
\end{table}}
}
% \vspace{-6mm}https://www.overleaf.com/project/60ad44361944029b2d77ecf3

\subsection{Problem Setup}
% Once we know the underlying causal generative process, it is easy to
% see transparently, the sources of spurious correlation or confounding.
% For example, an underlying disease can cause symptoms or
% manifest in their X-ray X in a patient. An additional variable U like
% the data-acquisition method (X-ray machine) between hospitals
% could act as a confounding variable, and affect both the diagnosis
% by doctors and symptoms or (X-ray quality) X. The associated
% causal model is then given by Fig. (1a). To train models that predict
% disease labels or risk scores based on patient symptoms or claims
% data X [35, 59]. Training a model that is free of such confounding
% then corresponds to specifying a distribution that will allow us
% to learn to predict labels by removing the effects of spurious
% correlation on the covariates X. This is where additional information
% of the graph and other covariates U, Z and D can help. The
% corresponding distribution we would ideally like to model is called
% an interventional distribution and graphically involves removing
% incoming edges to the variable we want to intervene on.
Once we know the underlying causal data-acquisition process, it is easy to transparently see the sources of spurious correlation which could act as confounding. For example, an underlying disease like pneumonia, $Y$  causes symptoms to manifest in a patient's X-ray, $\bX$. If the data-acquisition method, $\bU$ (e.g., X-ray machine) to acquire data $\bX$ (X-Rays) is correlated with the outcome $Y$ (presence of pneumonia), then $\bU$ acts as a confounding since it now affects both  
the symptoms $\bX$ (X-ray quality) and outcome $Y$. 
%This is to say if $\bU \rightarrow \bX$ exists and a spurious correlation adds the edge $\bU \rightarrow Y$, then we have $Y \leftarrow \bU \rightarrow \bX$, where $\bU$ is now a source of confounding bias (Fig. (\ref{fig:mes_con})).
This is shown by arrows $\bU \rightarrow \bX$ and $\bU \rightarrow Y$ creating a path $Y \leftarrow \bU \rightarrow \bX$. $\bU$ is called confounding bias, see Fig. (\ref{fig:mes_con}). 
To train models that predict disease label $Y$ based on patient symptoms $\bX$~\cite{zech2018variable, obermeyer2019dissecting}, one must mitigate direct effects from the confounding $\bU$ to the label $Y$ for which we will need \emph{interventional distributions.}
% Training a model to be free of this confounding requires specifying an \emph{interventional distribution} to learn to predict labels $Y$ free of the spurious correlation $\bU$.
% This graphically removes incoming edges to the variable (here, $Y$) that we want to \emph{intervene} on. Computing interventional distributions requires knowledge of the underlying acquisition process and some other variables (here, $\bU$, $\bZ$, $D$).
\paragraph{\textbf{Interventional distribution:}} An interventional distribution is one that is induced by local interventions on variables in a causal graph. The intervention breaks the causal relationship between the intervened variable and its parents. If we intervene on $Y$ and we're interested in characterizing the distribution over $\bX$, the corresponding interventional distribution is denoted by $P(\bx | do(Y=y))$. For e.g., in Fig.~\ref{fig:mes_con}, the value of $Y$ is fixed to some $y$ irrespective of the influence from $\bU$. 

Interventional distributions can be directly used to predict labels $Y$ free of the spurious correlation with $\bU$ as in \citet{subbaswamy2019preventing}. Here, for simple case of Fig.(\ref{fig:mes_con}): $P(y|\bx,do(y)) \propto P(\bx \mid do(y)) = \sum_u P(\bx|y,\bu)P(\bu)$, these models cannot be scaled to high-dimensional data. Instead,   
to learn a confounding free model, we sample from an interventional distribution. Since the goal is to learn labels $Y$ without being directly influenced by the confounding, $\bU$ (and unknown confounding), the desired interventional distribution we want samples from is denoted by $P(\bx|do(y))$.  Having samples from $P(\bx|do(y))$ instead of $P(\bx|y)$ ensures that the incoming edge from $\bU$ to $Y$ is broken, so that a deconfounded sample (i.e. $\bX, Y$ pairs) has no direct influence from the confounder $\bU$. Samples drawn from this interventional distributions are the de-confounded data (see Table~\ref{tab:terms}) that will be used to train an unbiased model trying to learn the conditional $P(y|\bx)$ ~\cite{pearl2011transportability}. 
However, this target interventional distribution is not always  \emph{identifiable}.
% in situations 
% This target interventional distribution is not always \emph{identifiable} 
% i.e. estimated from data collected from the original/un-intervened acquisition graph.    
% , theorem~\ref{thm:identifiability} provides an answer as to when this is possible.

\begin{theorem}[see~\cite{pearl2009causality}]\label{thm:identifiability}
For disjoint
variable sets $\mathbf{V}, \mathbf{W} \subseteq \mathbf{O}$ in a causal model, the effect of an intervention $do(\bv)$ on $\bV$ is said to be identifiable from the joint distribution over all variables $P(\bO)$ in the causal graph $\mathcal{G}$ over $\bO$, if $P(\bW|do(\bv))$ is (uniquely) computable from $P(\bO)$ in any causal model which induces $\mathcal{G}$.
\end{theorem}
A causal effect is identifiable, if such an expression can be found by applying the rules of do-calculus repeatedly. This follows directly from the definition of identifiability due to the fact that all observational distributions are assumed identical for the causal models that induce $\mathcal{G}$ \cite{pearl2009causality}. 
% The purpose of do-calculus is to represent the interventional distribution by using only observational probabilities. 
% In the causal graph $\mathcal{G}$ of interest, we first establish identifiability of the desired intervention using the identification rules of \citet{shpitser2006identification}. 
We apply rules of do-calculus in order to obtain an interventional distribution of interest~\citep{pearl1995causal}. The interventional distribution, if it exists, can be equivalently obtained using the \texttt{ID} algorithm~\citep{tian2003identification}. Applying these rules requires knowledge of the acquisition process $\cG$ and other variables (here, $\bU$, $\bZ$, $D$).

\paragraph{\textbf{do-calculus}}
We briefly review the rules of do-calculus. The purpose of do-calculus is to represent the interventional distribution $P(\mathbf{x}|do(y))$ using only observational probabilities. 
% A causal effect is identifiable, if such an expression can be found by applying the rules of do-calculus
% repeatedly. 
Let $\bX$, $\bY$ and $\bZ$ be pairwise disjoint sets of nodes in graph $\mathcal{G}$. Here $\mathcal{G}_{\overline{Y}, \underline{\mathbf{Z}}}$ means the graph that is obtained from $\mathcal{G}$ by removing all incoming edges to $Y$ and all outgoing edges of $Z$. Let $P$ be the joint distribution of all observed and unobserved variables. The following rules hold \cite{pearl1995causal}: 
\begin{enumerate}
    \item Insertion and deletion of observations:
 $$
P(\mathbf{x} \mid \mathbf{z}, \mathbf{w}, do(\mathbf{y}) )=P(\mathbf{x} \mid \mathbf{w}, do(\mathbf{y})), \text { if }(\mathbf{X} \perp \mathbf{Z} \mid \mathbf{Y}, \mathbf{W})_{\mathcal{G}_{\overline{\mathbf{Y}}}}
$$
\item Exchanging actions and observations: 
$$
P(\mathbf{x} \mid \mathbf{w}, do(\mathbf{y}), do(\mathbf{z}))= P(\mathbf{x} \mid \mathbf{z}, \mathbf{w}, do(\mathbf{y})), \text { if }(\mathbf{X} \perp \mathbf{Z} \mid \mathbf{Y}, \mathbf{W})_{\mathcal{G}_{\overline{\mathbf{Y}}, \underline{\mathbf{Z}}}}
$$
\item Insertion and deletion of actions: 
$$
P(\mathbf{x} \mid \mathbf{w}, do(\mathbf{y}), do(\mathbf{z}))= P(\mathbf{x} \mid \mathbf{w}, do(\mathbf{y})), \text { if }(\mathbf{X} \perp \mathbf{Z} \mid \mathbf{Y}, \mathbf{W})_{\mathcal{G}_{\overline{\mathbf{Y}}, \overline{ \mathbf{Z}(\mathbf{W})}}}
$$
where $
Z(\mathbf{W})=\mathbf{Z} \backslash An(\mathbf{W})_{\mathcal{G}_{\overline{\mathbf{Y}}}}
$
\end{enumerate}
% The proofs for the presented rules can be found in \cite{pearl1994probabilistic}. 
The expressions of all identifiable causal effects can be derived by using the three rules, implying  do-calculus is complete \cite{shpitser2006identification}.  

We now describe pretraining debiasing methods in the context of causal data-acquisition processes. 
% and the desired interventional distribution.
We compare commonly used methods that i) explicitly attempt to address the challenge of confounding bias to create \emph{de-confounded models} and ii) those that do not and instead create \emph{confounded models}.
\begin{asparaitem}
\item[] \emph{De-confounded models:}
Trained on data that is first de-confounded by some pre-training, e.g., Data-augmentation (DA), to proactively account for the confounding bias. This deconfounded data is subsequently used to train the model. These techniques will only use the data $\bX$ from de-confounded sample to predict $Y$.
\item[] \emph{Confounded models:} Directly trained on confounded data i.e. the original biased data. These correspond to methods that are either trained on all available data variables $\bX, \bZ, \bD, \bU$ or a subset thereof from the confounded data to predict $Y$.
\end{asparaitem}

\subsection{De-Confounding Methods}
\paragraph{\bf{Data augmentation (DA)}} 
In DA, we upsample in proportion to the training data for every specific confounding  \citep{panda2018contemplating,geirhos2018imagenet,zhang2020towards,goel2020model}. %But with bootstrapping, you just sample each instance with the weight we estimate. 
% For instance, deep networks for image tasks are commonly trained on a large number of rotated and translated versions of the original training data. 
While this is an effective method when the source of confounding is known, the number of samples required to re-balance \cite{pearl2011transportability} or building generative models \cite{geirhos2018imagenet} with pre-defined bias types can be prohibitive \cite{peyre2017weakly}.
When the source of confounding itself is unknown, simple data augmentation procedures cannot be used. %We use DA by utilising measured confounding ($\bU$) information in scenarios given by Fig.(\ref{fig:mes_con}), Fig.(\ref{fig:mes_wm}) and Fig.(\ref{fig:mes_wom}). 
%This method will only partially remove spurious correlations for data generated from  Fig.(\ref{fig:pmes_wm}) cannot be used for Fig.(\ref{fig:unmes_wm}). 
In DA, for label $Y$, number of samples for each value of $\bU$ is rebalanced by upsampling. Therefore, DA can successfully be applied for scenarios in
% will learn the desired interventional distribution $P(\bx|do(y))$ only for 
Fig.(\ref{fig:mes_con}), Fig.(\ref{fig:mes_wm}) and Fig.(\ref{fig:mes_wom}). It can only partially remove confounding for Fig.(\ref{fig:pmes_wm}) and cannot be applied to Fig.(\ref{fig:unmes_wm})
at all.   
% TODO: One sentence specific summary, E.g., We use DA with params ABC, set via XYZ, and optimize with 123.

\paragraph{\bf{Causal bootstrapping (CB)}}
Causal bootstrapping~\citep{little2019causal} is a sampling strategy that augments classical 
bootstrap re-sampling with casual information of the data acquisition process to generate samples from the %actual 
\emph{interventional distribution} we want to model. Any standard ML method can then be applied to this de-confounded data to train a de-confounded predictor. Thus, we can use out-of-the-box ML algorithms to train powerful, debiased models. We first derive the interventional distributions for the three cases in Fig. (\ref{fig:pmes_wm}), Fig. (\ref{fig:mes_wm}), and Fig.(\ref{fig:mes_wom}) not derived in \citet{little2019causal}:

\paragraph{Partially Observed Confounding with Mediator (Fig. \ref{fig:pmes_wm}):} To identify the $P(\mathbf{x}|do(y))$, we begin with factorization: 
\begin{equation}
   P(\mathbf{x}|do(y))= \sum_{u,z}P(\mathbf{x} \mid \mathbf{u},\mathbf{z}, do(y)) P(\mathbf{z}\mid do(y)) P(\mathbf{u} \mid do(y)) 
   \label{eq:fact_1}
\end{equation}
\begin{enumerate}
\item The term $P(\mathbf{x} \mid \mathbf{u},\mathbf{z}, do(y))$ in the sum is simplified by repeatedly using rule 2 and rule 3 of do calculus as: 
\begin{equation}
P(\mathbf{x} \mid \mathbf{u},\mathbf{z}, do(y)) 
= \sum_{y}P(\mathbf{x} \mid \mathbf{u}, y, \mathbf{z}) P(y \mid \mathbf{u}) 
\label{eq:tr_1_1}
\end{equation}
\vspace{-5mm}
\item The term $P(\mathbf{z} \mid do(y))$ in the sum is simplified using rule 2:
\begin{equation}
  P(\mathbf{z} \mid do(y)) = P(\mathbf{z} \mid y),         \text{       b.c }  (\mathbf{Z} \perp \mathbf{Y})_{\mathcal{G}_{\underline{Y}}}
  \label{eq:tr_2_1}
\end{equation}
\vspace{-5mm}
\item The term $P(\mathbf{u} \mid do(y))$ in the sum is simplified using rule 3: 
\begin{equation}
  P(\mathbf{u} \mid do(y)) = P(\mathbf{u}),         \text{       b.c }  (\mathbf{U} \perp Y)_{\mathcal{G}_{\overline{Y}}}
  \label{eq:tr_3_1}
\end{equation}
\vspace{-5mm}
\item Finally combining Eq. (\ref{eq:tr_1_1}), Eq. (\ref{eq:tr_2_1}) and Eq. (\ref{eq:tr_3_1}) into Eq. (\ref{eq:fact_1}):  
\begin{equation}
P(\mathbf{x}|do(y)) = \sum_{\bu,\bz}\left(\sum_{y'} P(\mathbf{x}|\bu,y',\bz) P(y'|\bu)\right)P(\bz|y)P(\bu)
\label{eq:intervention_1}
\end{equation}
\end{enumerate}

\paragraph{Observed Confounding with Mediator (Fig. \ref{fig:mes_wm}):} 
Similar to the previous scenario, we begin with the factorization: 
\begin{equation}
   P(\mathbf{x} \mid do(y))= \sum_{\bu,\bz}P(\mathbf{x} \mid \mathbf{u},\mathbf{z}, do(y)) P(\mathbf{z}\mid do(y)) P(\mathbf{u} \mid do(y)) 
   \label{eq:fact_2}
\end{equation}
\begin{enumerate}
\item The term $P(\mathbf{x} \mid \mathbf{u},\mathbf{z}, do(y))$ is simplified using rule 3: 
\begin{equation}
  P(\mathbf{x} \mid \mathbf{u},\mathbf{z}, do(y)) = P(\mathbf{x} \mid \mathbf{u},\mathbf{z}), 
\text{       b.c }  (\mathbf{X} \perp \mathbf{Y} \mid \bZ, \mathbf{U})_{\mathcal{G}} 
\label{eq:tr_1_2}
\end{equation}
\item $P(\mathbf{z} \mid do(y))$ is simplified using rule 2 similar to Eq.(\ref{eq:tr_2_1}).
% \begin{equation}
%   P(\mathbf{z} \mid do(y)) = P(\mathbf{z} \mid y),         \text{       b.c }  (\mathbf{Z} \perp \mathbf{Y})_{\mathcal{G}_{\underline{Y}}}
%   \label{eq:tr_2_2}
% \end{equation}
Similarly, $P(\mathbf{u} \mid do(y))$ is simplified using rule 3 similar to Eq.(\ref{eq:tr_3_1}).
% \begin{equation}
%   P(\mathbf{u} \mid do(y)) = P(\mathbf{u}),         \text{       b.c }  (\mathbf{U} \perp Y)_{\mathcal{G}_{\overline{Y}}}
%   \label{eq:tr_3_2}
% \end{equation}
Finally combining Eq. (\ref{eq:tr_1_2}), Eq. (\ref{eq:tr_2_1}) and Eq. (\ref{eq:tr_3_1}):
\begin{equation}
P(\mathbf{x} \mid do(y)) = \sum_{\bu,\bz}P(\mathbf{x} \mid \bu,\bz)P(\bz \mid y)P(\bu)
\label{eq:intervention_2}
\end{equation} 
\end{enumerate}

\paragraph{Biased care with observed confounding (Fig. \ref{fig:mes_wom}):}
Factorizing to obtain the conditional intervention distribution: 
\begin{equation}
   P(\mathbf{x} \mid do(y))= \sum_{\bu,\bd}P(\mathbf{x} \mid \mathbf{u}, do(y)) P(\mathbf{d}\mid \mathbf{u}, do(y)) P(\mathbf{u} \mid do(y)) 
   \label{eq:fact_3}
\end{equation}
\begin{enumerate}
\item The term $P(\mathbf{x} \mid \bu, do(y))$ in the sum is simplified using rule 2: 
\begin{equation}
    P(\mathbf{x} \mid  \bu, do(y)) = P(\mathbf{x} \mid \bu, y),  \text{       b.c }  (\mathbf{X} \perp Y)_{\mathcal{G}_{\underline{Y}}}
    \label{eq:tr_1_3}
\end{equation}
\vspace{-5mm}
\item The term $P(\mathbf{d} \mid \mathbf{u}, do(y))$ in the sum is simplified using rule 2:
\begin{equation}
  P(\mathbf{d} \mid  \bu, do(y)) = P(\mathbf{d} \mid \bu, y),  \text{       b.c }  (\mathbf{D} \perp Y)_{\mathcal{G}_{\underline{Y}}}
  \label{eq:tr_2_3}
\end{equation}
\vspace{-5mm}
\item $P(\mathbf{u} \mid do(y))$ is simplified using rule 3 similar to Eq.(\ref{eq:tr_3_1}).
% \begin{equation}
%   P(\mathbf{u} \mid do(y)) = P(\mathbf{u}),         \text{       b.c }  (\mathbf{U} \perp Y)_{\mathcal{G}_{\overline{Y}}}
%   \label{eq:tr_3_3}
% \end{equation}
Finally we combine Eq. (\ref{eq:tr_1_3}), Eq. (\ref{eq:tr_2_3}) and Eq. (\ref{eq:tr_3_1}) into Eq. (\ref{eq:fact_3}):  
\begin{equation}
P(\mathbf{x} \mid do(y)) = \sum_{\bu,\bd} P(\bx \mid \bu,y)P(\bd \mid \bu, y)P(\bu)
\label{eq:intervention_3}
\end{equation}
\end{enumerate}

\paragraph{{\textbf{Causal bootstrap weights:}}} The approach involves expressing the interventional distribution as a simple weighted KDE to generate sampling weights, $w_n(c)$ for $n^{th}$ sample and class $y=c$:
\begin{equation}
    P(\mathbf{x} \mid do(y = c)) \approx \sum_{n \in N} K\left[\mathbf{x}-\mathbf{x}_{n}\right] w_{n}(c)
    \label{eq:cau_boo_fun}
\end{equation}
Where $\bx_n$ corresponds to $n^{th}$ data sample from the observational data with $N$ samples. 
Therefore the weights for the different causal graphs in Fig.(\ref{fig:pmes_wm}), Fig.(\ref{fig:mes_wm}) and Fig.(\ref{fig:mes_wom}) are:

\paragraph{Partially observed confounding with mediator (Fig.\ref{fig:pmes_wm})}
The interventional distribution in Eq. (\ref{eq:intervention_1}) can be written as 
\begin{equation}
P(\mathbf{x}|do(y = c)) = \sum_{\bu,\bz}\left(\sum_{y'} P(\mathbf{x},\bu,\bz, y')\frac{P(y'/\bu) P(\bz/y=c) P(\bu)}{P\left(\bu, y', \bz\right)}\right)
\label{eq:intervention_1_joint}
\end{equation}
Joint distribution can be estimated non-parametrically using KDE: 
\begin{equation}
    P(\mathbf{x},\bu,\bz,y') \approx \frac{1}{N} \sum_{n \in N} K\left[\mathbf{x}- \mathbf{x_{n}}\right]K\left[\mathbf{u}- \mathbf{u_{n}}\right]K\left[\mathbf{z}- \mathbf{z_{n}}\right]K\left[\mathbf{y'}- \mathbf{y'_{n}}\right] 
    \label{eq:joint_kde_1}
\end{equation}
Inserting Eq. (\ref{eq:joint_kde_1}) into Eq. (\ref{eq:intervention_1_joint}) we get: 
% \begin{equation}
\begin{multline*}
% P(\mathbf{x}|do(y)) \approx \frac{1}{N}\sum_{\bu,\bz,y}\left(\sum_{n \in N} K\left[\mathbf{x}- \mathbf{x_{n}}\right]K\left[\mathbf{u}- \mathbf{u_{n}}\right]K\left[\mathbf{z}- \mathbf{z_{n}}\right] \\K\left[\mathbf{y'}- \mathbf{y'_{n}}\right] \frac{P(y'/\bu) P(\bz/y) P(\bu)}{P\left(\bu, y', \bz\right)}\right)
P(\mathbf{x}|do(y = c)) \approx \sum_{n \in N} K\left[\mathbf{x}- \mathbf{x_{n}}\right]  \frac{1}{N}\sum_{\bu,\bz,y'}K\left[\mathbf{u}- \mathbf{u_{n}}\right]K\left[\mathbf{z}- \mathbf{z_{n}}\right] \\ K\left[\mathbf{y'}- \mathbf{y'_{n}}\right]\frac{P(y'/\bu) P(\bz/y=c) P(\bu)}{P\left(\bu, y', \bz\right)}
\end{multline*}
% \label{eq:intervention_1_joint_KDE}
% \end{equation}
Expressing the desired intervention $P(\bx|do(y))$ in the form Eq.(\ref{eq:cau_boo_fun}), 
% can be expressed as a weighted KDE:
% $$P(\mathbf{x}|do(y)) \approx \sum_{n \in N} K\left[\mathbf{x}- \mathbf{x_{n}}\right]w_n$$
the desired weights $w_n(c)$ are given as 
{\small$$w_n(c) = \left(\frac{1}{N} \sum_{\bu,\bz,y'}K\left[\mathbf{u}- \mathbf{u_{n}}\right]K\left[\mathbf{z}- \mathbf{z_{n}}\right]K\left[\mathbf{y'}- \mathbf{y'_{n}}\right] 
\frac{P(y'/\bu) P(\bz/y=c) P(\bu)}{P\left(\bu, y',\bz\right)}\right)$$}
% $$ w_n = \left( \frac{1}{N}\sum_{\bu,\bz,y'}K\left[\mathbf{u}- \mathbf{u_{n}}\right]K\left[\mathbf{z}- \mathbf{z_{n}}\right]K\left[\mathbf{y'}- \mathbf{y'_{n}}\right]  \frac{ P(\bz/y)}{P\left(\bz/ \bu, y' \right)}\right) $$
After further factorizing and simplifying: 
% $$ w_n = \frac{1}{N}\sum_{\bu,\bz}K\left[\mathbf{u}- \mathbf{u_{n}}\right]K\left[\mathbf{z}- \mathbf{z_{n}}\right]\left(\sum_{y'} K\left[\mathbf{y'}-\mathbf{y'_{n}}\right]  \frac{ P(\bz/y)}{P\left(\bz/ \bu, y' \right)}\right) $$
$$ w_n(c) = \frac{1}{N}\sum_{\bu,\bz}K\left[\mathbf{u}- \mathbf{u_{n}}\right]K\left[\mathbf{z}- \mathbf{z_{n}}\right]  \left.\frac{P(\bz/y=c)}{P\left(\bz/ \bu, y' \right)}\right|_{y'=y'_{n}} $$
Every instance of variable $y'$ 
is replaced with it's realization $y'_n$ . 
Similarly factorize for $\{\bu, \bz\}$. 
% \begin{equation}
%   w_n = \frac{\hat{P}\left(\bz_{n} \mid y\right)}{N \hat{P}\left(\bz_{n} \mid y_{n},\bu_{n}\right)} 
%   \label{wei:pmes_0}
% \end{equation}
% Since we are dealing with discrete sample space for $y \in \Omega_{Y} = \{c_1, c_2, \dots, c_k\}$, for each class $y=c$, the weights are given as: 
% \begin{equation}
%   w_n = \frac{\hat{P}\left(\bz_{n} \mid y = c\right)}{N \hat{P}\left(\bz_{n} \mid y_{n},\bu_{n}\right)} 
%   \label{wei:pmes}
% \end{equation}

\paragraph{Observed confounding with mediator (Fig.\ref{fig:mes_wm})}
The interventional distribution in Eq. (\ref{eq:intervention_2}) can be re-written as: 
\begin{equation}
P(\mathbf{x}|do(y = c)) = \sum_{\bu,\bz}\left( P(\mathbf{x},\bu,\bz)\frac{P(\bz/y=c) P(\bu)}{P\left(\bu,\bz\right)}\right)
\label{eq:intervention_2_joint}
\end{equation}
Following a similar procedure we obtain weights $w_n(c)$: 
$$ w_n(c) = \left( \frac{1}{N}\sum_{\bu,\bz}K\left[\mathbf{u}- \mathbf{u_{n}}\right]K\left[\mathbf{z}- \mathbf{z_{n}}\right]\frac{ P(\bz/y=c)}{P\left(\bz/ \bu \right)}\right) $$

\paragraph{Biased care with observed confounding (Fig.\ref{fig:mes_wom}):}
The interventional distribution in Eq. (\ref{eq:intervention_3}) can be re-written as: 
\begin{equation}
P(\mathbf{x}|do(y = c)) = \sum_{\bu,\bd}\left( P(\mathbf{x},\bu,y=c)\frac{P(\bd/\bu, y=c) P(\bu)}{P\left(\bu, y=c\right)}\right)
\label{eq:intervention_3_joint}
\end{equation}
Estimating the joint by KDE and expressing the desired intervention $P(\bx|do(y))$ in the form Eq.(\ref{eq:cau_boo_fun}), we obtain weights $w_n(c)$: 
$$ w_n(c) = \left( \frac{1}{N}\sum_{\bu,\bd}K\left[\mathbf{u}- \mathbf{u_{n}}\right]I\left[y_{n} = c\right]\frac{ P(\bd/\bu,y = c)}{P\left(y = c/ \bu \right)}\right) $$
For a classification task with discrete sample space for $y \in \Omega_{Y} = \{c_1, c_2, \dots, c_K\}$  
, given causal graph $\mathcal{G}$ from Fig. \ref{fig:DAG}, the weights $w_n(c)$ for a given class $y = c$ are given by:
\begin{equation}
w_{n}(c)=\left\{\begin{array}{ll}
\frac{I\left[y_{n}=c\right]}{N P\left(y=c \mid \bu_{n}\right)}, & \text { if } \mathcal{G} \text { is Fig }(\ref{fig:mes_con}) \\
\frac{P\left(\bz_{n} \mid y=c\right)}{N P\left(\bz_{n} \mid \bu_{n} \right)}, & \text { if } \mathcal{G} \text {is Fig }(\ref{fig:mes_wm}) \\
\frac{P\left(\bz_{n} \mid y=c\right)}{N P\left(\bz_{n} \mid y_{n}, \bu_{n} \right)}, & \text { if } \mathcal{G} \text {is Fig }(\ref{fig:pmes_wm}) \\
\frac{P\left(\bz_{n} \mid y=c\right)}{N P\left(\bz_{n} \mid y_{n}\right)}, & \text { if } \mathcal{G} \text { is Fig }(\ref{fig:unmes_wm}) \\
\frac{\sum_{d} I\left[y_{n}=c\right]P\left(\bd_{n} \mid y=c, \bu_{n}\right)}{N P\left(y = c \mid \bu_{n}\right)}, & \text { if } \mathcal{G} \text { is Fig }(\ref{fig:mes_wom})
\end{array}\right.
\label{eg:weights}
\end{equation}
For each class $y=c$, we can now re-sample $\bx$ from the observed distribution using kernel function $K$ centered on $\bx_n$, with probability $w_n(c)$ obtained by the desired interventional distribution $P(\bx|do(y) = c)$. This provides de-confounded samples which are simulated from observational data. For simple classification problems, kernel $K$ can just be a delta function, making $\bx$ = $\bx_n$. In this case we sample $\bx_n$ with probability $w_n(c)$ for class $c$.
% A detailed derivation of the interventional distribution and the bootstrap weights for all cases in Fig.~\ref{fig:DAG} are provided in Appendix \ref{sec:appendix}.
This causal method allows us to model the associated conditional distribution devoid of confounding biases, without explicitly collecting data from the interventional distribution. We apply causal bootstrapping using confounding ($\bU$), mediator ($\bZ$) and level of care ($D$) information, as given by weights in Eq. \ref{eg:weights} to scenarios in Fig. (\ref{fig:DAG}). The overall procedure to estimate the weights and sample deconfounded data is provided in Algorithm~\ref{algo:train}. The Algorithm takes as input the `confounded data' sample $\bX_{conf}, Y_{conf}$ and any additional covariates as available $\bU, \bZ, D$, and graph $\cG$. First, we check if $P(\bx|do(y))$ is identifiable using the \texttt{ID} algorithm or do-calculus rules. We cannot truly debias a model if $P(\bx|do(y))$ is not identifiable (using any technique). If the distribution is identifiable, we proceed with an iterative procedure to determine sample weights per label in $\Omega_Y$ to generate $X_{deconf}, Y_{deconf}$. We can use $X_{deconf}, Y_{deconf}$ with any standard ML method and no longer need $\bU, \bZ$ and $D$ to train our model. Therefore the information from $\bU, \bZ$ and $D$ is only used prior to training and is not required for prediction. Thus acquisition cost of confounding and mediator information is a one time cost.% only.% and not required at test time.    
\begin{algorithm}[htbp!]
\begin{footnotesize}
\begin{algorithmic}[1]
%\SetAlgoLined
%\DontPrintSemicolon
\State \textbf{Input:} $\mathcal{G}$, $\bX_{conf} \in \Omega_{X}, Y_{conf} \in \Omega_{Y} = \{c_{1},c_{2}, \dots c_{K}\}$,  
\State \textbf{Output:} $ \bX_{deconf}, Y_{deconf}$ 
\State \textbf{Additional CB inputs:}  $\bZ_{conf}$, $\bD_{conf}$, and/or $\bU_{conf}$, as observed.  
% \If{$P (\mathbf{x} \mid do( y))$ is identifiable from $\mathcal{G}$}
\State Find interventional distribution $P (\mathbf{x} \mid do( y))$ using do-calculus.  
    \If{$P (\mathbf{x} \mid do( y))$ is identifiable from $\mathcal{G}$}
    \For{$y = c$} 
        \For {$n \in N$}
        \State Compute $w_n$ from Eq. \eqref{eg:weights}
        \EndFor
    \State Sample $\lfloor N p(y = c)\rfloor$ samples from                   $\Omega_{\bX}$, sampling $\bx_n$ with                probability $w_n(c)$
    \State For sampled $\bx_n$, set $y_n$ = c
    \EndFor
\Else
    \State FAIL
\EndIf
\end{algorithmic}
\end{footnotesize}
\caption{\small{Causal bootstrapping for De-biasing data for classification for graphs $\mathcal{G}$ in Fig. (\ref{fig:DAG})}}
\label{algo:train}
%\vspace{1mm}
\end{algorithm}

% \paragraph{{\bf{Confounding Methods}}}
\subsection{Confounding Methods}
We use the term "confounding methods" to refer to methods that rely on original (confounded) data.
\paragraph{\bf{Informative features (IF)}} Common knowledge in deep learning advocates using as many features as possible for prediction, to allow deep networks to learn separable representations from raw features \citep{suresh2017clinical,ghassemi2014unfolding}. 
% This method also does not guarantee unbiased models especially if the source of confounding is unobserved.
For scenarios in Fig. (\ref{fig:mes_con}) and Fig. (\ref{fig:mes_wom}) where we have measured confounding, confounding ($\bU$) is used along with image ($\bX$) to train a model. For scenarios in Fig. (\ref{fig:mes_wm}), Fig. (\ref{fig:pmes_wm}) and Fig. (\ref{fig:unmes_wm}) we use both mediator ($\bZ$) and confounding ($\bU$) information along with image ($\bX$) to train the models. A challenge with IF method is that mediator (eg., bio-markers) and confounding information is required at test time which can be prohibitively expensive.% making this a costly process. 

% \paragraph{Simple} \textcolor{red}{Some methods (Simple) train directly on confounded data only using features that they want the model to learn their label based on. This could mean consciously excluding sensitive information which could act as source of bias; for example using patient claims data to predict risk scores while consciously excluding patients' race \citep{obermeyer2019dissecting} or excluding information that is ideally uninformative of the label; for example using patient chest x-ray data to predict disease labels while not accounting for hospital metadata \citep{zech2018variable}.} 

\paragraph{\bf{Simple}} Another common practice is to only use manually selected features ($\bX$) from the original confounded data, that we want the model to learn their label based on. We call this method `Simple', and train models using only $\bX$ for all data scenarios. 

\label{sec:data}
\section{Experimental Setup}
Data generation-acquisition mechanisms can be arbitrarily complex. Here, we use five fundamental scenarios that capture commonly observed confounding biases in practice. These scenarios are analogous to causal models in Fig. (\ref{fig:DAG}). A method that succeeds in these scenarios is a strong debiasing pre-training method.

\subsection{Acquisition Scenarios}
We motivate each of the five cases using real-world 
examples. Consider data that is used to train models to predict disease labels 
$Y$ based on patient symptoms or X-rays ($\bX$). 
Data-acquisition methods across hospitals represented by $\bU$ could act as a confounding variable, and affect both the disease outcomes, $Y$ and symptoms, $\bX$. Therefore, 
our confounding is either a spurious correlation or information that we do not want our model to learn based on. For e.g., hospital specific information for predicting disease label based on medical imaging data, patients' race for risk prediction using claims data 
are examples of  
confounding~\citep{zech2018variable,obermeyer2019dissecting}. 

\paragraph{\textbf{Observed confounding:}} In some cases, confounding ($\bU$) is observed and available for model training. % when the model is being trained. 
For e.g., hospital specific information or patient's race could be recorded as in Fig. (\ref{fig:mes_con}).
% \texttt{A} . 

\paragraph{\textbf{Observed confounding with mediator:}} A mediating variable directly influences a patient's X-ray, $\bX$, based on patients' underlying disease, $Y$ and is affected by confounding (X-ray machine specifications) $\bU$ only through disease $Y$. For instance, in \citet{zech2018variable} a disease bio-marker $\bZ$ could be unaffected by $\bU$ given label $Y$, and therefore act as a mediating variable. Mediators can help in settings where precise confounding variables are not known i.e. if X-ray machine specs are not recorded as in Fig. (\ref{fig:mes_wm}). 

\paragraph{\textbf{Partially observed confounding with mediator:}} In most practical scenarios, the type of confounding is unknown and identifying all confounding is impossible. For instance, the US Affordable Care Act requires collecting race information but not patient's socio-economic condition which could act as an unobserved confounding in addition to the recorded race $\bU$~\citep{national2016metrics,nuru2018relative}. In such situations, the presence of mediating variable $\bZ$ can be beneficially used to learn unbiased models using the graph in Fig.~(\ref{fig:pmes_wm}).

\paragraph{\textbf{Unobserved confounding with mediator:}} During data observation or collection, the presence of possible confoundings could be unknown and hence unobserved. Even if all confounders are known, they may be unavailable (missing) or not collected because obtaining such information comes at a high cost to patient confidentiality as well as to the institution. In the above examples, if we are unaware of race or hospital information being a confounding signal or are unable to collect this data, we have a hidden confounding. In the most general setting, such confounding cannot be corrected for~\citep{shpitser2008complete} as the target interventional distribution is unidentifiable. However, 
this scenario too stands to gain from the availability of a mediator $\bZ$. Hence we focus our analysis on completely hidden confounding with mediators (Fig.~(\ref{fig:unmes_wm})). 

\paragraph{\textbf{Biased care with Observed confounding:}} In some cases, the level of care given to a patient could be influenced by the confounding bias i.e. unwarranted reliance on race in prioritizing patients for medical care~\citep{obermeyer2019dissecting}. Accounting for this bias with knowledge of the care level $D$ can help debias models (Fig.~(\ref{fig:mes_wom})).

\subsection{Evaluation Methodology}\label{sec:target_env}
To quantify whether a specific procedure can truly learn unbiased models, we use four separate test settings for evaluation. None of the methods have access to these test samples during training or validation. 

\paragraph{\textbf{Confounded (Conf) test}} 
This test data is generated with the same distribution 
as the training set. Testing on this data exclusively can be misleading, and is a standard practice in ML. Performance on this sample in relation to other evaluations will tell us if the model is learning on biases.

\paragraph{\textbf{Unconfounded (Unconf) test}}
This data consists of samples free of any confounding bias. 
Performance on this data reveals how reliant a model actually is on confounding bias for training.

\paragraph{\textbf{Reverse-confounded (Rev-Conf) test}} 
% This setting reverses the confounding relationship by switching the bias to a different class from the training set. 
This setting creates a shift in the confounding-label distribution by reversing the correlations found in the training set i.e, If in the training set most positive disease labels ($Y$ =1) come from hospital $U=1$ and controls ($Y$=0) come from hospital $U=0$, then it's vice-versa in the test set.
%We will hereon refer to these as reverse-confounded data. 
Good performance on this set confirms that the debiasing method has learned an unbiased model. Deteriorated performance indicates model relies on the confounding. % as it is able to generalize on data with significant covariate shift.

\paragraph{\textbf{Unseen test}} 
This setting consists of unconfounded data with different confounding biases i.e, If the training data has data from hospitals (biases) $U=\{0,1\}$ then the unseen test set contains data from hospital $U=2$. This measures the ability of the model to generalize to novel biases unseen during training or validation.

If a model has learnt to predict based on the confounding in the dataset, it will perform well on \emph{Confounded test} set but %fail to do so for 
 not on other test sets. However, a model not relying on confounding bias will perform similarly across \emph{all} test cases, with the exception of the Unseen test, where domain shifts could lead to decreases in model performance regardless.
 \begin{figure*}[htbp!]
    \centering
    \includegraphics[width=.85\linewidth, trim=0 30 0 20, clip ]{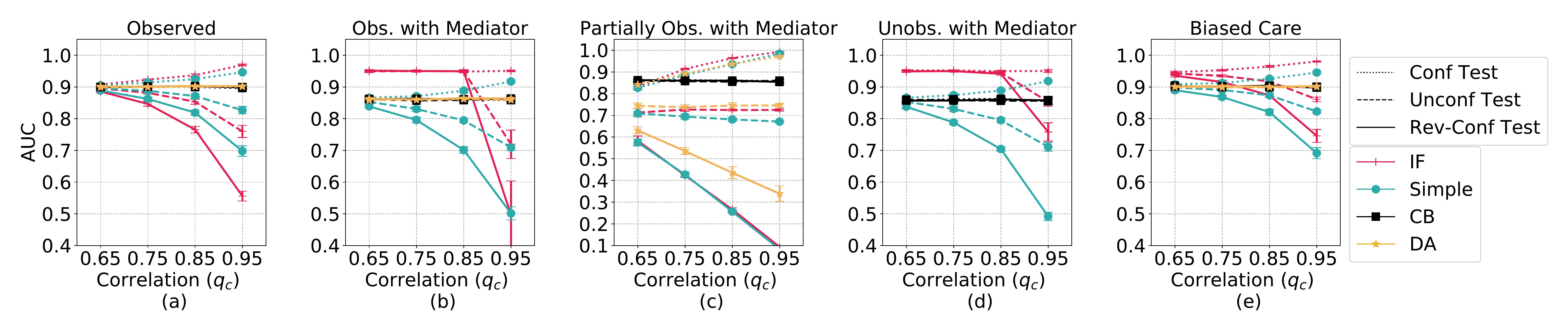}
    \caption{Performance of confounded (IF and Simple) and de-confounded (CB and DA) models on synthetic data for different levels of correlation.   
    The confounded model performance on \emph{Unconf} and \emph{Rev-conf} test data decreases significantly at higher correlations showing these models are biased.}
    \label{fig:dataset_corr}
\end{figure*}

\begin{figure*}[htbp!]
    \centering
    \begin{minipage}[t]{0.55\linewidth}
    \includegraphics[width=\linewidth, trim=0 0 250 0, clip ]{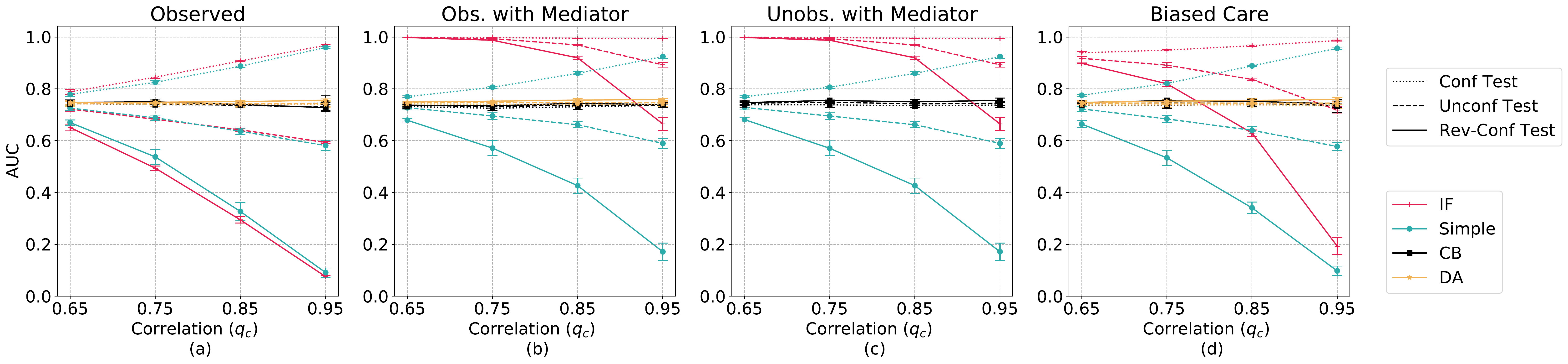}
    \caption{Performance of confounded (IF and Simple) and de-confounded (CB and DA) models learnt on MIMIC-CXP data with real shifts for different levels of correlation. We do not study MIMIC-CXP on partially obs. with mediator scenario (Fig. \ref{fig:pmes_wm}) to avoid introducing synthetic confounding. The de-confounded model performance on \emph{Unconf} and \emph{Rev-conf} test remain constant across scenarios, showing these models are unbiased. Legend shared with Fig.~\ref{fig:dataset_corr}.
    }
    \label{fig:dataset_corr_NIH_MIMIC}
    \end{minipage}
    \hfill
    \begin{minipage}[t]{0.43\linewidth}
    \centering
    \includegraphics[width=0.7\linewidth,]{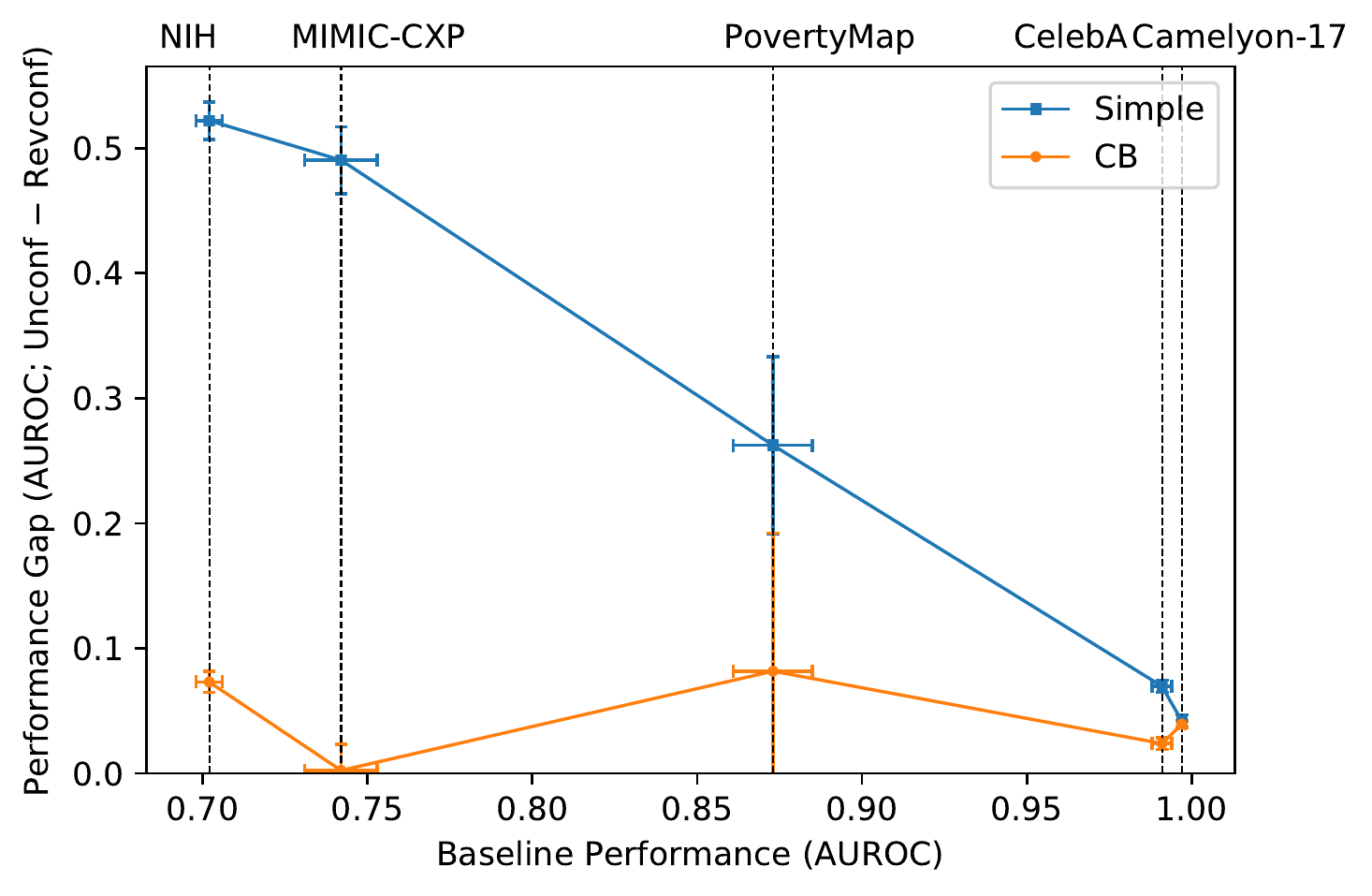}
    \caption{
    Comparison between the performance drops of a) ``Simple'' (confounded model) b) ``CB'' (un-confounded models) on predictive tasks of decreasing complexity trained on datasets with $q_c$ = 0.95 for ``Observed confounding (Fig.(\ref{fig:mes_con}))'' scenario. We observe that the performance gap (i.e. the level of reliance on confounding bias) decreases with increasing baseline performance for the ``Simple'' model.}
    \label{fig:data_comp}
    \end{minipage}
\end{figure*}

\label{sec:experiments}
\subsection{Data Simulation}

Quantifying failure modes of 
deep learning methods requires that we precisely control the level of confounding. We therefore create synthetic and semi-synthetic datasets for the five acquisition settings from Fig.~\ref{fig:DAG}.
% A detailed description of data generation process that respects the causal graphs for all datasets in provided in the Appendix \ref{sec:appendix}.
Sampling from any causal graph $\mathcal{G}$ in Fig.(\ref{fig:DAG}) can be performed using the following conditional distributions as applicable for the causal graph of interest:
%\begin{equation*}
 \begin{alignat*}{5}
    Y & \sim \text {Bernoulli}(p), %\\
    && U &&\mid &&Y  & &\sim \text {Bernoulli}\left(q(y)\right)\\
    Z \mid Y & \sim \text {Bernoulli}\left(r(y)\right)\ %\\
     D &&\mid &&Y,&&U  &&\sim \text {Bernoulli}\left(f(y,u)\right)%\\
 \end{alignat*}
%\end{equation*} 
For the partially observed confounding case in Fig. (\ref{fig:pmes_wm}), the unobserved confounding $V$ is modelled as:
\begin{equation*}
    \begin{aligned}
    V \mid Y & \sim \text {Bernoulli}\left(q'(y)\right)\\
    \end{aligned}
\end{equation*}
This is used only for data-generation. 
%Note that we can always find the equivalent conditional $Y|U$ using Bayes’ rule, since the factorized version of the joint distribution remains the same. The main difference is that marginal estimates are pre-determined, making it easier to adjust to label imbalanced datasets. 
Parameter $p$ denotes the probability with which $Y=1$. While, 
Parameter $q(y)$ models the relation between the $Y$ and confounding $U$. For different settings of correlation, if $q(1)$ = $q_{c}$, then $q(0)$ = $1-q_{c}$. 
% Here $q_{c}$ denotes the chance with which the sample is confounded $U=1$, given outcome label $Y=1$ i.e, $p(U=1|Y=1)$. 
Wherever applicable, the mediator $Z$ has conditional parameter for $r(0) = p(Z=1 \vert Y=0) = 0.05$ and $r(1)= p(Z=1 \vert Y=1) = 0.95$. The level of care label $D$ is generated using the conditional distribution $f(1,0) = 1 - f(0,0) = p(D=1 \vert Y=1,U=0)= 0.8$ and $f(1,1) = 1 - f(0,1) = p(D=1 \vert Y=1,U=1)= 0.95$.

\subsection{Datasets}\label{sec:datasets}

\paragraph{{\bf{Synthetic Data:} }}
    For synthetic data generation, $\bX$ is modeled as a mixture of Gaussians: 
    %\begin{align*}
     $\bX \mid Pa(\bX)  \sim \mathcal{N}\left(\mu(Pa(\bX)), \sigma^{2}\right)$. 
    %\end{align*}
    Here $Pa(\bX)$ refers to the parents of $\bX$ in the causal graph $\mathcal{G}$. 
    
\paragraph{{\bf{Semi-synthetic Data}}} We sample and modify the following imaging datasets: \begin{inparaenum} \item[a)] CelebA for gender classification \citep{liu2015faceattributes}, \item[b)] ChestXray8 (NIH) for atelectasis classification \citep{wang2017chestx}, \item[c)] MIMIC-CXR for atelectasis classification \citep{johnson2019mimic}, \item[d)] CheXpert for atelectasis classification \cite{irvin2019chexpert}, \item[e)] Camelyon17 for tumor prediction \cite{koh2020wilds}, \item[f)] PoveryMap for classification of asset index above median~\cite{yeh2020using}\end{inparaenum}. 
%using the above settings. % for different prediction tasks. %Where image, $\bX$ is sampled as: 
    % \begin{align*}
    %     \boldsymbol{X} \mid Pa(\bX) & \sim \operatorname{IMAGE}(Pa(\bX))
    % \end{align*}
Depending on the causal graph $\mathcal{G}$, we sample an image $\bX$ based on the label (and mediating information) if they are parents $Pa(\bX) = \{Y, Z\}$ in $\cG$ and then transform the image by inducing random confounding according to $\bU$ and $\bV$ (as applicable). $V$ is used only for data generation.

{\vspace{-2mm}
\small{
\begin{table}[!h]
    \centering
    \resizebox{.99\linewidth}{!}{
 \begin{tabular}{c||c|c||c} 
 \hline
  \textbf{Dataset} & \textbf{U=1} & \textbf{U=0} & \textbf{Unseen Domain}\\ 
 \hline\hline
%   MNIST & Rotate/Trans. & No Transform\\
%  \hline
  CelebA & Rotate $90\degree$ & No Transform & N/A\\
 \hline
  NIH & Rotate $90\degree$ & No Transform & N/A\\ 
 \hline 
  Camelyon17 & Hospital 3 & Hospital 4 & Hospital 5\\
 \hline 
 PovertyMap & Malawi, Tanzania & Kenya, Nigeria & 19 Other Countries\\
 \hline
  MIMIC-CXP & MIMIC-CXR & CheXpert & NIH\\ [1ex] 
 \hline
\end{tabular}
}
\caption{Summary of semi-synthetic datasets with synthetic and real-world confounding ($\bU$). Other variables of interest - mediator ($\bZ$) and care level ($D$) are synthesized acc. to respective distributions. The domain of the Unseen test is shown for datasets where an external domain is available.}
\label{tab:datasets}
\vspace{-8mm}
\end{table}}}
\subsection{Inducing Confounding}
Confounding is any effect observed in the data $\bX$ that we do not want to rely on to train our model. For each evaluation scenario, we generate: \begin{inparaenum} \item[i)] the confounded training data \item[ii)] the evaluation test sets described in Sec~\ref{sec:target_env}\end{inparaenum}. Table~\ref{tab:datasets} summarizes our data generation procedure for all experiments.

\paragraph{{\bf{Synthesized Shifts:}}}
We use rotation of the image to act as confounding ~\cite{rabanser2019failing}. 
An image is rotated by fixed $\theta = 90\degree$ counter-clockwise. We use this synthetic shift to induce confounding in NIH and CelebA. 
% For each type (as appropriate), we explore different levels of confounding intensity to understand the effect of confounding: 
% \begin{asparaenum}
% \item[a)] \textbf{Rotation} : An image is rotated by fixed $\theta \in \{30, \, 50, \, 70, \, 90\}$ in the counter clockwise direction.      
% \item[b)] \textbf{Translation} : An image is translated by fixed image shift of  $\delta \times \, \text{image-width}$, where $\delta \in \{0.3,\, 0.5,\, 0.7,\, 0.9\}$
% \end{asparaenum}
\paragraph{{\bf{Real Shifts:}}}
We also evaluate methods on real-world shifts ($\bU \rightarrow \bX$). The WILDS dataset provides in-the-wild distribution shifts for diverse data modalities and applications~\cite{koh2020wilds}. We use 2 datasets from WILDS and construct an X-ray dataset MIMIC-CXP:
\begin{asparaenum}
\item[a)]  {\bf{Camelyon17}}: Here hospital of acquisition is treated as a natural source of confounding $\bU$. 
\item[b)] {\bf{PovertyMap:}} We treat the country of acquisition as a natural source of confounding $\bU$.
%\end{inparaitem}
% where the task is to predict the presence of tumor tissues in a given image. 
% \item[b)] 
% From within the WILDS datasets, we choose another dataset - PovertyMap dataset.
% % where the task is to predict the presence of tumor tissues in a given image. 
% We consider the country of acquisition as a natural source of confounding $\bU$. 
\item[c)] {\bf{MIMIC-CXP}}: Constructed by sampling from two popular chest x-ray datasets - MIMIC-CXR \cite{johnson2019mimic}, collected at the Beth Israel Deaconess Medical Center in Boston, and CheXpert \cite{irvin2019chexpert}, collected at the Stanford Hospital. Here, the hospital of acquisition is designed to be the natural source of confounding $\bU$.
\end{asparaenum} %A pretraining debiasing method that works well in spite of such shifts is a strong debiasing method. 

\label{sec:results}

\section{Results}
\subsection{Performance Across Methods and Acquisition Scenarios}
\paragraph{\textbf{Motivation}}
We investigate the extent to which each method in Sec \ref{sec:methods_all} helps to learn unbiased models. Biased models should have poorer performance on unconfounded and reverse-confounded test data than on confounded test. A successfully debiased model will perform well for all test data. We measure the AUC of models learnt using Simple, IF, DA, and CB methods on all test sets.  

\paragraph{\textbf{Experiment}}
In Fig.~(\ref{fig:dataset_corr}) and Fig.~(\ref{fig:dataset_corr_NIH_MIMIC}) we compare model performance on i) synthetic data and ii) semi-synthetic MIMIC-CXP data with real world shifts, at different confounding levels $q_c$ for all test cases (see Sec.\ref{sec:target_env}), for all evaluation scenarios. The X-axis shows increasing level of spurious correlation $q_c \in \{0.65,0.75,0.85,0.95 \}$ between the confounding and label. In Table~\ref{tab:deconf_comp}, we showcase one demonstrative example comparing performance of methods on MIMIC-CXP data with very strong spurious correlation $q_c = 0.95$. 
\paragraph{\textbf{Results}}
\paragraph{Confounding Methods lead to biased models}
In Fig.(\ref{fig:dataset_corr}) and Fig.(\ref{fig:dataset_corr_NIH_MIMIC}) we notice that ``Simple'' models perform well on data drawn from Conf-test, but deteriorate with a large margin on Unconf and Rev-conf $(>41.9\%)$ test sets (for $q_c=0.95$).
For scenarios without mediator information, i.e. ``Observed'' (Fig.\ref{fig:mes_con}) and ``Biased care''  (Fig.\ref{fig:mes_wom}), IF model performances deteriorate with increasing confounding correlations, similar to ``Simple'' models $(>22.0\%)$ (for $q_c=0.95$). In cases with mediator information ``Obs. with Mediator''(Fig.\ref{fig:mes_wm}), ``Partially Obs. with Mediator''(Fig. ~\ref{fig:pmes_wm}) and ``Unobs. with Mediator''(Fig. ~\ref{fig:unmes_wm}), IF training is robust for low correlations but fails with a large margin for higher confounding correlations.
IF model performance drop on Rev-conf test shows their reliance on confounding.% bias. 

\paragraph{De-Confounding Methods lead to unbiased models}
De-confounding methods, CB and DA  perform similarly with little difference across all correlations in scenarios when the source of confounding is known - ``Observed''(Fig.~\ref{fig:mes_con}), ``Obs. with Mediator'' (Fig.~\ref{fig:mes_wm}) and ``Biased Care''(Fig.~\ref{fig:mes_wom}). However, DA is only applicable when the confounding is known and measured, limiting its utility. When the source of confounding is only partially known in ``Partially Obs. with Mediator'' (Fig.~\ref{fig:pmes_wm}), DA can train only partially unbiased models. As shown, on synthetic data in Fig.(\ref{fig:dataset_corr})(c), DA performance is significantly lower on Unconf and Rev-Conf test ($>24\%$) showing its dramatic failure in cases when confounding is only partially known. However, hidden and partially observed confounding is a more practically occurring non-trivial scenario. In contrast, CB models leverage mediators and can adjust for all potential unobserved confounding and shows similar performance across the ``Partially Obs. with Mediator'' (Fig.\ref{fig:pmes_wm}) and ``Unobs. with Mediator'' (Fig.\ref{fig:unmes_wm}).   
Table~\ref{tab:deconf_comp} shows similar performances across all scenarios in the most correlated case $q_c = 0.95$.
We find that CB is very effective even for cases with ``Unobs. with Mediator'' (Fig.\ref{fig:unmes_wm}). CB needs mediator information only during pre-training, hence models trained with CB can be deployed with no mediator information. These results show that CB is highly beneficial for learning in all data acquisition scenarios, establishing the benefit of using causal knowledge for pre-training debiasing.

\begin{table*}[!htbp]
\resizebox{0.8\textwidth}{!}{\begin{tabular}{l|l|l|l|l|l}
\textbf{Setting}             & \textbf{Test Data} & \textbf{DA}   & \textbf{IF}   & \textbf{Simple} & \textbf{CB}   \\ \hline
\multirow{2}{*}{\textbf{Obs.  U (\ref{fig:mes_con})}} & \textbf{Unconf}    & 
\multirow{2}{*}{\hspace{-2.1mm}$\left.\begin{array}{l}
                0.743 \pm 0.011\\
                0.756 \pm 0.005\\
                \end{array}\right\rbrace \textbf{\textcolor{green}{0.013}}$} & 
\multirow{2}{*}{\hspace{-2.1mm}$\left.\begin{array}{l}
                0.593 \pm 0.003\\
                0.075 \pm 0.004\\
                \end{array}\right\rbrace \textbf{\textcolor{red}{-0.518}}$} &
\multirow{2}{*}{\hspace{-2.1mm}$\left.\begin{array}{l}
                0.581 \pm 0.020\\
                0.091 \pm 0.018\\
                \end{array}\right\rbrace \textbf{\textcolor{red}{-0.490}}$} &
\multirow{2}{*}{\hspace{-2.1mm}$\left.\begin{array}{l}
                0.730 \pm 0.017\\
                0.728 \pm 0.012\\
                \end{array}\right\rbrace \textbf{\textcolor{red}{-0.002}}$}                
                 \\
                             & \textbf{Reverse}   &  & &    &  \\
                             & \textbf{Unseen}      & $\boldsymbol{0.746 \pm 0.005} $ & N/A           & $0.643 \pm 0.036$   & $\boldsymbol{0.740 \pm 0.012}$ \\ \hline
                             
\multirow{3}{*}{\textbf{Obs.  U with Z (\ref{fig:mes_wm})}} & \textbf{Unconf}    & \multirow{2}{*}{\hspace{-2.1mm}$\left.\begin{array}{l}
                0.748 \pm 0.004\\
                0.758 \pm 0.004\\
                \end{array}\right\rbrace \textbf{\textcolor{green}{0.010}}$} & 
\multirow{2}{*}{\hspace{-2.1mm}$\left.\begin{array}{l}
                0.893 \pm 0.009\\
                0.665 \pm 0.025\\
                \end{array}\right\rbrace \textbf{\textcolor{red}{-0.228}}$} &
\multirow{2}{*}{\hspace{-2.1mm}$\left.\begin{array}{l}
                0.590 \pm 0.019\\
                0.171 \pm 0.034\\
                \end{array}\right\rbrace \textbf{\textcolor{red}{-0.419}}$} &
\multirow{2}{*}{\hspace{-2.1mm}$\left.\begin{array}{l}
                0.745 \pm 0.008\\
                0.756 \pm 0.009\\
                \end{array}\right\rbrace \textbf{\textcolor{green}{0.011}}$}                
                 \\
                             & \textbf{Reverse}   &  & &    &  \\
                             & \textbf{Unseen}      & $\boldsymbol{0.749 \pm 0.006} $ & N/A           & $0.654 \pm 0.044$   & $\boldsymbol{0.743 \pm 0.008}$ \\ \hline
                             
\multirow{3}{*}{\textbf{Unobs.  U with Z (\ref{fig:unmes_wm})}} & \textbf{Unconf}    & N/A           & \multirow{2}{*}{\hspace{-2.1mm}$\left.\begin{array}{l}
                0.893 \pm 0.009\\
               0.665 \pm 0.025\\
                \end{array}\right\rbrace \textbf{\textcolor{red}{-0.228}}$} & 
                
      \multirow{2}{*}{\hspace{-2.1mm}$\left.\begin{array}{l}
               0.590 \pm 0.019\\
               0.171 \pm 0.034\\
                \end{array}\right\rbrace \textbf{\textcolor{red}{-0.419}}$}   &
    \multirow{2}{*}{\hspace{-2.1mm}$\left.\begin{array}{l}
               0.736 \pm 0.005\\
               0.739 \pm 0.010\\
                \end{array}\right\rbrace \textbf{\textcolor{green}{0.003}}$} \\
                             & \textbf{Reverse}   & N/A           &  &   & \\
                             & \textbf{Unseen}      & N/A           & N/A           & $0.654 \pm 0.044$   & $\boldsymbol{0.740 \pm 0.008}$ \\ \hline

\multirow{3}{*}{\textbf{\begin{tabular}[l]{@{}l@{}}Obs. U with\\ biased care (\ref{fig:mes_wom})\end{tabular}}} & \textbf{Unconf}  &  \multirow{2}{*}{\hspace{-2.1mm}$\left.\begin{array}{l}
                0.745 \pm 0.005\\
                0.759 \pm 0.008\\
                \end{array}\right\rbrace \textbf{\textcolor{green}{0.014}}$} & 
\multirow{2}{*}{\hspace{-2.1mm}$\left.\begin{array}{l}
                0.718 \pm 0.015\\
                0.193 \pm 0.033\\
                \end{array}\right\rbrace \textbf{\textcolor{red}{-0.525}}$} &
\multirow{2}{*}{\hspace{-2.1mm}$\left.\begin{array}{l}
                0.578 \pm 0.016\\
                0.097 \pm 0.019\\
                \end{array}\right\rbrace \textbf{\textcolor{red}{-0.481}}$} &
\multirow{2}{*}{\hspace{-2.1mm}$\left.\begin{array}{l}
                0.736 \pm 0.006\\
                0.742 \pm 0.015\\
                \end{array}\right\rbrace \textbf{\textcolor{green}{0.006}}$}                
                 \\
                             & \textbf{Reverse}   &  & &    &  \\
                             & \textbf{Unseen}      & $\boldsymbol{0.746 \pm 0.006} $ & N/A           & $0.636 \pm 0.034$   & $\boldsymbol{0.740 \pm 0.003}$ \\ \hline
\end{tabular}
}
\caption{AUC difference of confounded (methods: IF and Simple) and de-confounded models (methods: CB and DA) learnt on the MIMIC-CXP dataset with $q_c=0.95$ and hospital location as confounding. Performance of CB models across scenarios shows the effectiveness of causal bootstrap de-confounding to learn without relying on confounding bias. The DA models perform comparably well, though they cannot be applied when the confounding is not fully observed.}
\label{tab:deconf_comp}
\end{table*}

\subsection{Performance Across Tasks}
\paragraph{\textbf{Motivation}} 
We analyse how performance of deep networks trained using ``CB'' and the naive ``Simple'' method varies across task complexity. We define task complexity as the strength of the invariant correlation, i.e, the ease with which a model is able to learn the association $\bX \rightarrow Y$ in the absence of spurious confounding. We vary this by varying the datasets shown in Table (\ref{tab:datasets}). Note that we empirically find the strength of $\bX \rightarrow U$ to be strong in all cases.

\paragraph{\textbf{Experiment}} 
In Figure (\ref{fig:data_comp}) we examine the most spuriously confounded scenarios by fixing $q_c = 0.95$ for all datasets. Results for other levels of confounding show a similar trend, although with less significant drops at lower correlations. The X-axis shows the baseline performance i.e, training and testing on unconfounded data, which shows strength of the association $\bX \rightarrow Y$. The Y-axis is the AUROC difference between the \emph{Unconf} test - \emph{Rev-conf} test which shows model dependence on confounding bias. A larger gap suggests that the model is more biased.
\begin{figure*}[htbp!]
    \centering
    \includegraphics[width=0.85\linewidth, trim=0 0 0 0, clip ]{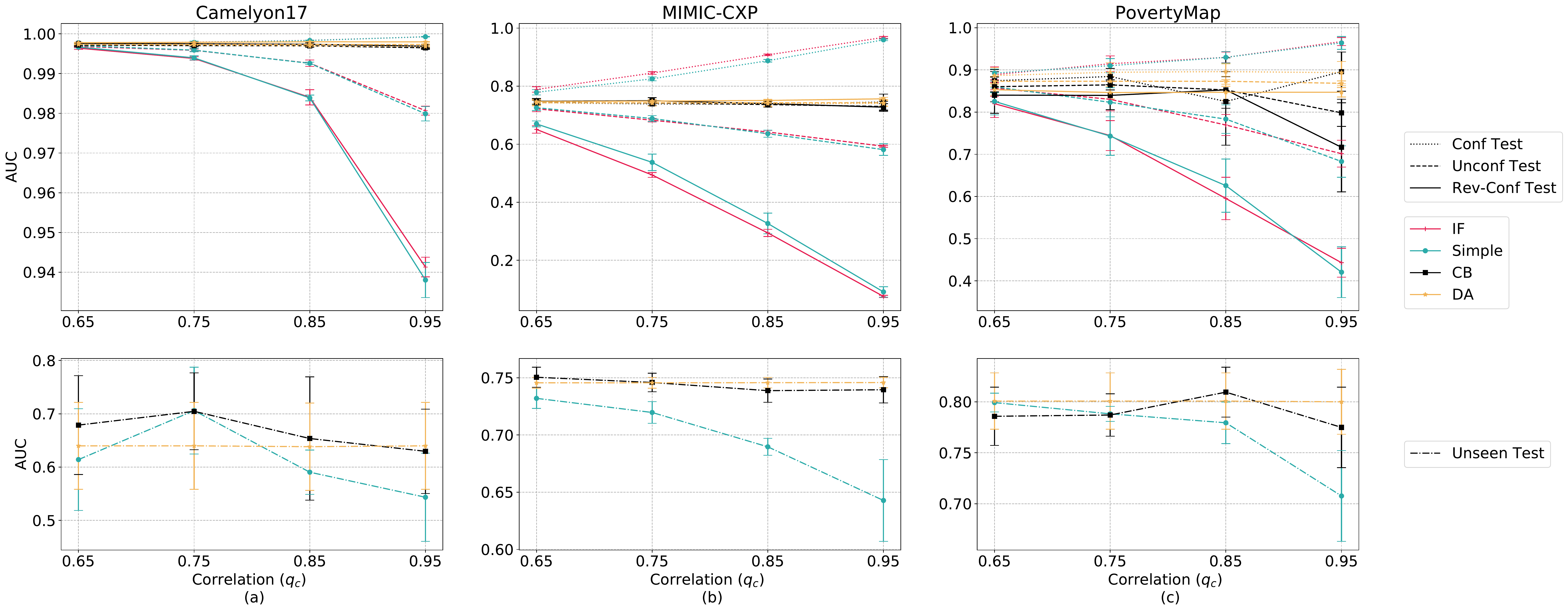}
    \caption{Performance of confounded (methods: IF and Simple) and de-confounded (methods: CB and DA) models learnt on a) Camelyon dataset b) MIMIC-CXP dataset and c) PovertyMap dataset with real shifts for different levels of correlation.
    (bottom row) 
    The confounded models' performance on unseen test sets  decreases significantly at higher correlations showing these models are biased. The deconfounded models' performance on unseen test sets shows they are learning invariant associations and are unbiased.}
    \label{fig:dataset_corr_real}
\end{figure*}

\paragraph{\textbf{Results}. Complex tasks more susceptible to bias} 
We find that confounded models learnt using the ``Simple'' method show decreasing performance gap with decreasing task complexity. Thus, ``Simple'' models are more prone to confounding for tasks where the invariant correlation is weaker. CB models however show similar performance gaps throughout, implying that they do not depend on the spurious correlation regardless of task complexity.         

\subsection{Performance on Real World Shifts}
\paragraph{\textbf{Motivation}} 
We analyze the effectiveness of CB in de-confounding data with real-world shifts. We compare against other methods highlighting its ability to learn unbiased models and hence generalize better to completely unseen biases. 
% domains. 
% If the methods have helped learn models that have learnt on invariant associations rather than on spurious confoundings, they should be able to perform well on test data from a different domain.    
\paragraph{\textbf{Experiment}} 
In Fig.~(\ref{fig:dataset_corr_real}) we compare model performance on Camelyon17, NIH-MIMIC and PovertyMap datasets with real world shifts ($\bU \rightarrow \bX$) at different confounding levels $q_c$ for all test cases (see Sec.\ref{sec:target_env}). We show this analysis only on the observed confounding scenario in Fig.(\ref{fig:mes_con}) where each variable $\bX$, $Y$ and $\bU$ is observed in the datasets and no variable is synthesized. Additionally, we also show the performance of the methods on test data from unseen biases (here domains) as listed in Table \ref{tab:datasets}.
% \begin{inparaenum}
%     \item For Camelyon dataset - test images are from hospital 5.
%     \item For MIMIC - CXR dataset - test data is from NIH-Chestxray18. 
%     \item For PovertyMap dataset - test data is from all $19$ countries not present in training data. 
% \end{inparaenum}
\paragraph{\textbf{Results}. CB helps learn invariant associations}
We notice that across the three datasets, ``Simple'' models perform well on data drawn from \emph{Conf} test, but deteriorate with a large margin on \emph{Unconf} and \emph{Rev-Conf} test sets. Deconfounding methods on the other hand show similar performance across correlations and test environments. We see similar behaviour in their performances on test data from different domains implying that CB is indeed helping models learn invariant associations. 

\label{sec:discussion}
\section{Discussion}
Training ML models that are robust to spurious correlations is especially beneficial for safety-critical applications. Here we systematically investigate benefits of pre-training methods designed to train unbiased models. Using five complex but practical confounded generation-acquisition  scenarios, we conclude that commonly used practices in current ML are insufficient to learn truly robust deep models. Our generative scenarios are designed to cover fundamental but practical scenarios reflecting data-collection practices or well known disparities in healthcare\footnote{\url{https://www.healthypeople.gov/2020/about/foundation-health-measures/Disparities}}~\citep{chen2020ethical}. For medical imaging tasks we see drops in AUC of up to $50\%$ across all scenarios. 
% Methodologically, we significantly extend Causal Bootstrapping, a causal debiasing procedure to these complex acquisition scenarios. 
% and is the only method to uniformly help train unbiased models.
 %Failure of other methods suggests significant work is required to improve pre-training pipelines to learn unbiased models. 
We show that CB  
accounts for domain knowledge using causal mechanisms and is applicable across all acquisition scenarios as long as required interventional distribution is \emph{identifiable}.
% considered here. \textcolor{red}{Further, the method can be applied to more general acquisition scenarios too } 
This highlights the need to incorporate causal view along with domain knowledge when building reliable deep models. 
% We conclude that  knowledge of data-collection processes by understanding the causal relationships and identifiability is essential to characterizing when and if such unbiased models can be trained.
Our investigation is complementary to methods attempting to build robustness to confounding during training~\cite{janizek2020adversarial,subbaswamy2019preventing} or those using multiple environments to capture the desired invariances~\cite{arjovsky2019invariant,heinze2018invariant,peters2015causal}. %In many cases, eliciting exact shifts across environments which some methods rely on~\citep{subbaswamy2019preventing} is challenging and a 
Many such methods focus on worst-case adversarial robustness~\citep{janizek2020adversarial} which can decrease utility~\citep{hu2018does}. %Therefore pre-processing methods without such information is highly desirable. 
Our analyses are a first investigative step towards a causal view to design pre-training methods that could easily be adopted by practitioners.
%For instance, Fig.~(\ref{fig:unmes_wm}) and~(\ref{fig:pmes_wm}) reflect scenarios where expensive bio-marker information is available during training, but not for a new incoming patient. Similarly Fig.~(\ref{fig:mes_wom}) reflects a scenario where existing disparate treatment practices in healthcare suggest disparate care for racial groups~\citep{obermeyer2019dissecting}.
 % In the future we will extend this analysis to in-training methods as well as other complex domains like text, increasingly used in safety-critical applications.

\begin{acks}
We would like to thank Taylor Killian for his valuable feedback. Resources used in preparing this research were provided, in part, by the Province of Ontario, the Government of Canada through CIFAR, and companies sponsoring the Vector Institute. Dr. Marzyeh Ghassemi is funded in part by Microsoft Research, a Canadian CIFAR AI Chair held at the Vector Institute, a Tier 2 Canada Research Council Chair, and an NSERC Discovery Grant. 
\end{acks}

\bibliographystyle{ACM-Reference-Format}
\bibliography{acmart}

\begin{appendices}
\label{sec:appendix}

\end{appendices}

\end{document}